\documentclass[10pt,twocolumn,letterpaper]{article}

\usepackage[pagenumbers]{cvpr} 

\usepackage{graphicx}
\usepackage{amsmath}
\usepackage{amssymb}
\usepackage{booktabs}
\usepackage{microtype}
\usepackage[font={footnotesize}]{caption}
\usepackage{enumitem}
\usepackage{array}
\usepackage{flushend}
\usepackage{cuted}
\usepackage{placeins}

\usepackage[export]{adjustbox}
\usepackage{etoolbox}

\usepackage{xcolor}
\definecolor{dark-green}{RGB}{12,80,12}
\definecolor{red}{RGB}{255,0,0}
\newcommand{\secref}[1]{Sec.~\ref{#1}}
\newcommand{\figref}[1]{Fig.~\ref{#1}}

\newcommand{\rot}[1]{\rotatebox[origin=c]{90}{#1}}
\newcolumntype{P}[1]{>{\centering\arraybackslash}p{#1}}
\usepackage[pagebackref,breaklinks,colorlinks]{hyperref}
\usepackage{mathtools}

\newcommand{\tabref}[1]{Tab.~\ref{#1}}
\usepackage[capitalize]{cleveref}
\crefname{section}{Sec.}{Secs.}
\Crefname{section}{Section}{Sections}
\Crefname{table}{Table}{Tables}
\crefname{table}{Tab.}{Tabs.}

\def\cvprPaperID{10801} 
\def\confName{CVPR}
\def\confYear{2022}

\begin{document}

\title{Amodal Panoptic Segmentation}

\author{Rohit Mohan
\qquad
Abhinav Valada\\
University of Freiburg\\
\tt\small\{mohan, valada\}@cs.uni-freiburg.de}

\maketitle

\begin{abstract}
   Humans have the remarkable ability to perceive objects as a whole, even when parts of them are occluded. This ability of amodal perception forms the basis of our perceptual and cognitive understanding of our world. To enable robots to reason with this capability, we formulate and propose a novel task that we name amodal panoptic segmentation. The goal of this task is to simultaneously predict the pixel-wise semantic segmentation labels of the visible regions of stuff classes and the instance segmentation labels of both the visible and occluded regions of thing classes. To facilitate research on this new task, we extend two established benchmark datasets with pixel-level amodal panoptic segmentation labels that we make publicly available as KITTI-360-APS and BDD100K-APS. We present several strong baselines, along with the amodal panoptic quality (APQ) and amodal parsing coverage (APC) metrics to quantify the performance in an interpretable manner. Furthermore, we propose the novel amodal panoptic segmentation network (APSNet), as a first step towards addressing this task by explicitly modeling the complex relationships between the occluders and occludes. Extensive experimental evaluations demonstrate that APSNet achieves state-of-the-art performance on both benchmarks and more importantly exemplifies the utility of amodal recognition. The benchmarks are available at \url{http://amodal-panoptic.cs.uni-freiburg.de}.
\end{abstract}

\section{Introduction}\label{sec:intro}

\begin{figure}
    \centering
    \begin{subfigure}[b]{\linewidth}
        \centering
        \includegraphics[trim=0 0 0 140,clip,width=\textwidth]{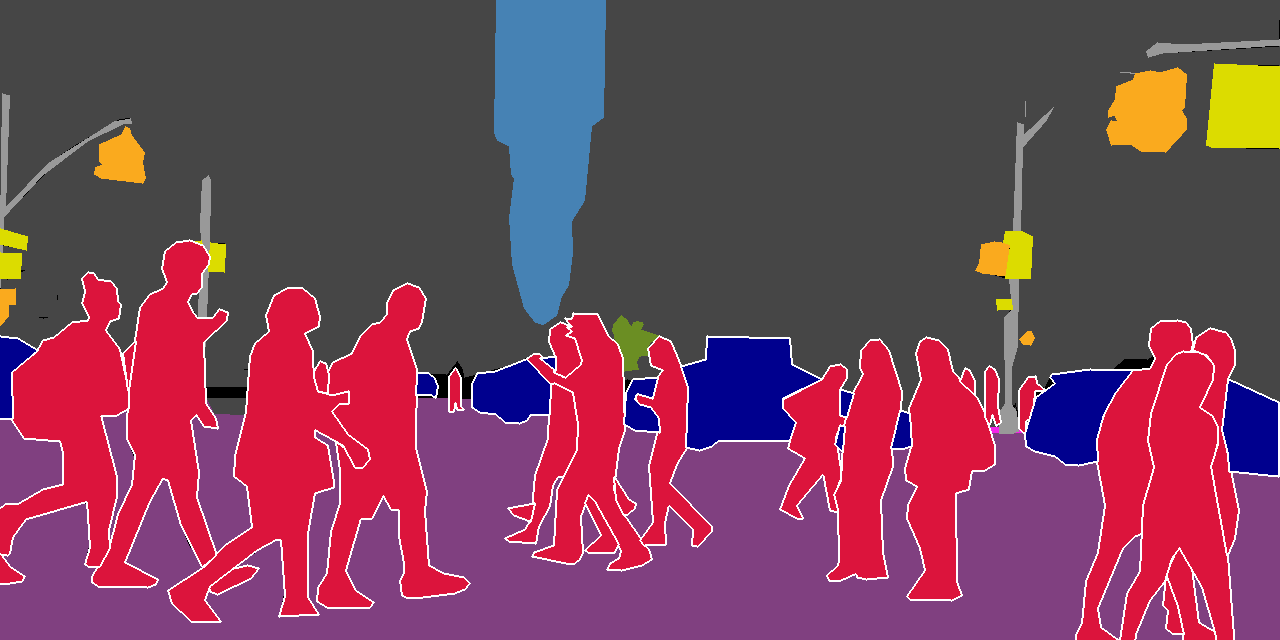}
        \subcaption{Panoptic Segmentation}
        \label{panoptic_segmentation_eg}
    \end{subfigure}
    \begin{subfigure}[b]{\linewidth}
        \centering 
        \includegraphics[trim=0 0 0 140,clip,width=\textwidth]{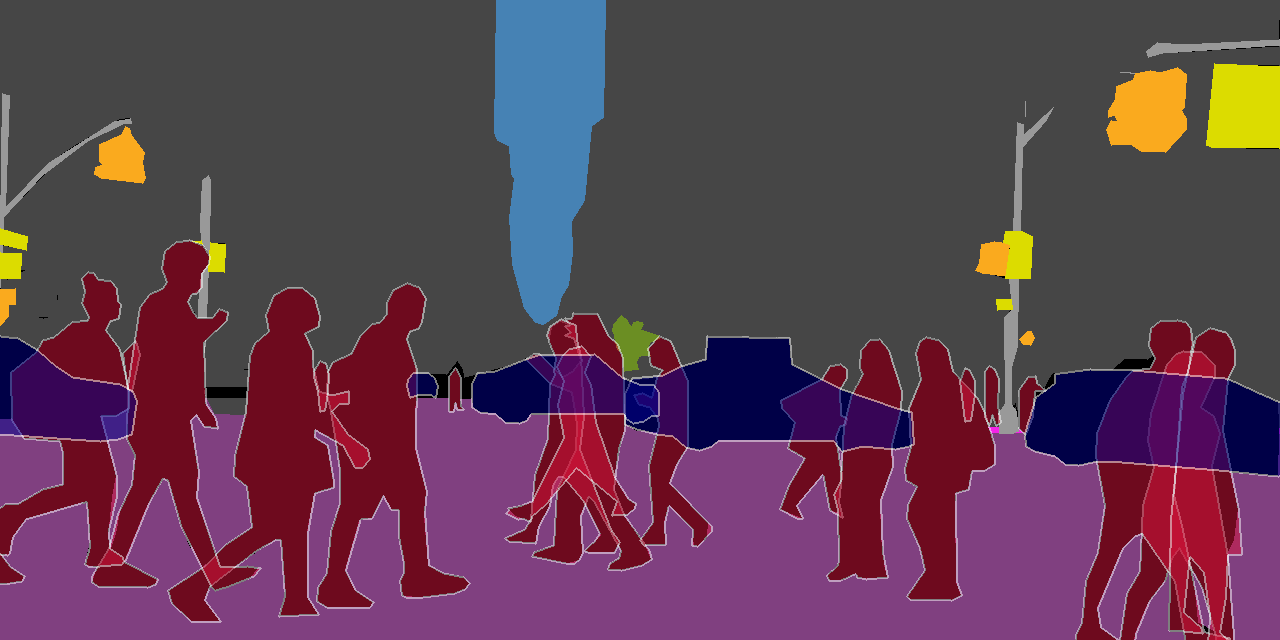}
        \subcaption{Amodal Panoptic Segmentation}
        \label{amodal_panoptic_segmentation_eg}
    \end{subfigure}
    \vspace{-0.5cm}
    \caption{Illustration of \subref{panoptic_segmentation_eg} panoptic segmentation and \subref{amodal_panoptic_segmentation_eg} amodal panoptic segmentation that encompasses visible regions of \textit{stuff} classes, and both visible and occluded regions of \textit{thing} classes as amodal masks.}
    \label{fig:paper-teaser}
    \vspace{-0.5cm}
\end{figure}

Humans rely on their ability to perceive complete physical structures of objects even when they are only partially visible, to navigate through their daily lives~\cite{nanay2018importance}. This ability, known as amodal perception, serves as the link that connects our perception of the world to its cognitive understanding. However, unlike humans, robots are limited to modal perception~\cite{valada2016convoluted, zurn2020self, mittal2018vision}, which restricts their ability to emulate the visual experience that humans have. In this work, we bridge this gap by proposing the amodal panoptic segmentation task.\looseness=-1

Any given scene can broadly be categorized into two components: \textit{stuff} and \textit{thing}. Regions that are amorphous or uncountable belong to \textit{stuff} classes (e.g., sky, road, sidewalk, etc.), and the countable objects of the scene belong to \textit{thing} classes (e.g., cars, trucks, pedestrians, etc.). The amodal panoptic segmentation task illustrated in \figref{fig:paper-teaser}~\subref{amodal_panoptic_segmentation_eg} aims to concurrently predict the pixel-wise semantic segmentation labels of visible regions of \textit{stuff} classes, and instance segmentation labels of both the visible and occluded regions of \textit{thing} classes. We believe this task is the ultimate frontier of visual recognition and will immensely benefit the robotics community. For example, in automated driving, perceiving the whole structure of traffic participants at all times, irrespective of partial occlusions~\cite{valverde2021there}, will minimize the risk of accidents. Moreover, by inferring the relative depth ordering of objects in a scene, robots can make complex decisions such as in which direction to move relative to the object of interest~\cite{hurtado2021learning} to obtain a clearer view without additional sensor feedback.\looseness=-1

Amodal panoptic segmentation is substantially more challenging as it entails all the challenges of its modal counterpart (scale variations, illumination changes, cluttered background, etc.) while simultaneously requiring more complex occlusion reasoning. This becomes even more complex for non-rigid classes such as pedestrians. These aspects also reflect on the groundtruth annotation effort that it necessitates. In essence, this task requires an approach to fully grasp the structure of objects and how they interact with other objects in the scene to be able to segment occluded regions even for cases that seem ambiguous.

Our contributions in this paper are twofold. First, we propose the novel task of amodal panoptic segmentation, a comprehensive scene recognition problem. To fully establish the task as well as to encourage future research, we extend two challenging urban driving datasets with amodal panoptic segmentation labels to create the KITTI-360-APS and BDD100K-APS benchmarks. We present several baselines for this task by combining state-of-the-art amodal instance segmentation methods with top-down panoptic segmentation networks. Further, we introduce two evaluation metrics referred to as amodal panoptic quality (APQ) and amodal parsing coverage (APC), to coherently quantify the performance of segmentation of \textit{stuff} classes in visible regions and \textit{thing} classes in both visible and occluded object regions. The APQ metric measures the performance independent of the size of instances and the APC metric considers the size of instances while giving more importance to the segmentation quality of larger objects than smaller objects. We introduce the size-dependent metric since a variety of applications seek high-quality segmentation of objects closer to the camera than far away objects, such as in autonomous driving.

Second, we propose the novel \mbox{APSNet} architecture that consists of a shared backbone and task-specific semantic and amodal instance segmentation heads followed by a parameter-free fusion module that yields the amodal panoptic segmentation output. In our approach, we split the amodal bounding box contents into the visible region mask of the target object, the occluded region mask of the target object referred to as the occlusion mask, and the object masks that occludes the target object referred to as the occluder. The occluder and occlusion features enable the amodal mask head to identify occlusion regions, while the visual and occlusion features enable the network to predict the amodal mask of the object. Furthermore, we refine the visible mask with amodal features in conjunction with visible features to impart occlusion awareness. To prevent the loss of localization of features in favor of semantic features, we increase the receptive field for context aggregation with dilated convolutions instead of downsampling in the semantic head. We make our code and models publicly available at \url{http://amodal-panoptic.cs.uni-freiburg.de}.


\section{Related Work}
Panoptic segmentation approaches can be categorized into proposal-free and proposal-based methods. Proposal-free methods~\cite{Gao_2019_ICCV, cheng2020panoptic, wang2020axial} first perform semantic segmentation, followed by applying various techniques to group \textit{thing} pixels such as instance center regression~\cite{uhrig2018box2pix}, Hough-voting~\cite{leibe2004combined}, or pixel affinity~\cite{keuper2015efficient} to obtain instance segmentation. On the other hand, in proposal-based methods~\cite{gosala2022bird, mohan2020efficientps, sirohi2021efficientlps}, typically a network head generates the object bounding boxes along with their masks and a parallel head yields the semantic segmentation output. In this work, we propose a top-down amodal panoptic segmentation architecture. We choose the top-down over the bottom-up approach due to its ability to handle large-scale variation in instances which plays a vital role in segmenting \textit{thing} class objects.

Li~\textit{et~al.}~\cite{li2016amodal} introduce the amodal instance segmentation task for which their approach relies on the directions of high heatmap values computed for each object to iteratively enlarge the corresponding object modal bounding box. Follmann~\textit{et.~al}~\cite{follmann2019learning} propose a class-specific amodal instance segmentation approach called ORCNN which replaces the single instance mask head of Mask R-CNN~\cite{he2017mask} with amodal and inmodal instance mask heads. Further, they employ an occlusion mask prediction head on top of the modal-specific heads. Subsequently, Qi~\textit{et~al.}~\cite{qi2019amodal} introduce the multi-level coding module to explicitly impart global information for better segmentation of the occluded area. VQ-VAE~\cite{jang2020learning} replaces the fully convolutional instance mask heads with variational autoencoders. Their method first classifies the input features into intermediate shape codes and then recovers complete object shapes from the intermediate shape codes. To learn the aforementioned discrete shape codes, they pretrain a vector quantized variational autoencoder model on the amodal groundtruth masks. Xiao~\textit{et~al.}~\cite{yuting2021amodal} use a shape-prior memory codebook with an autoencoder to refine the initial amodal mask prediction from Mask R-CNN. Similar to~\cite{jang2020learning}, they pretrain the autoencoder on amodal groundtruth masks. More recently, BCNet~\cite{Ke_2021_CVPR} employs two overlapping GCN layers that detect the occluding objects and partially occluded object instances to decouple the boundaries of both the occluding and occluded object instances.

Lastly, Zhu~\textit{et al.}~\cite{zhu2017semantic} propose amodal semantic segmentation with the COCO amodal dataset. Their task requires the prediction of visible and invisible regions of \textit{thing} classes in a class-agnostic manner while allowing multiple detections of the same objects. The COCO amodal dataset does not provide any labels for amorphous regions (wall, floor, etc.). In contrast, we introduce two benchmark datasets that treat all prominent amorphous regions (road, sidewalk, etc.) and non-traffic participants (pole, fence, etc.) in an urban scene as \textit{stuff}, similar to the standard convention followed in panoptic segmentation~\cite{kirillov2019panoptic}. Consequently, our datasets consider all the traffic participants (cars, pedestrians, etc.) as part of \textit{thing} classes. Furthermore, our amodal panoptic segmentation task allows at most one semantic label and instance-ID assignment to the pixel of visible regions. This discourages overlaps and requires predictions to be class-specific.

\section{The Amodal Panoptic Segmentation Task}
For a given set of $C$ semantic classes, the goal of the amodal panoptic segmentation task is to map each pixel $i$ of a given input image to a set $A_i$ comprising pairs of $(c, \kappa, v_) \in C \times N \times V$, where $c$ represents the semantic class for the pixel, $\kappa$ represents the instance ID, and $v \in V$ represents the visibility of the prediction pair where $V$ is encoded as $V \in \{1,2\}$. Here, $\kappa$ of each pair in set $A_i$ associates a group of pixels that have the same semantic class but belong to a different segment, and are unique for each segment for the given image. $v$ determines whether the corresponding $\kappa$ is the visible part ($v=1$) of its segment or the occluded part ($v=2$). Moreover, in the set $A_i$, at most one pair with $v=1$ is feasible. Additionally, for $ c \in C_{s}$ the corresponding $\kappa$ is irrelevant, where $C_s$ is the subset of $C$ that consists of \textit{stuff} semantic classes.  

For simplicity, we can define the amodal panoptic segmentation task at the object segment level. Given an input image, the task aims to predict all the visible \textit{stuff} class segments where each \textit{stuff} class can have at most one segment associated with it. In contrast, for \textit{thing} classes, each class can have more than one visible segment associated with it. Further, the segmentation of each \textit{thing} class segment can comprise both visible and occluded region segmentation.

\section{Evaluation Metrics}
In this section, we present the metrics that we use to evaluate the performance of amodal panoptic segmentation. 

\subsection{Amodal Panoptic Quality}

In order to facilitate quantitative evaluation, we adapt the standard panoptic quality (PQ)~\cite{kirillov2019panoptic} metric used for quantifying the performance of panoptic segmentation by accounting for the invisible or occluding regions in our new metric that we name amodal panoptic quality (APQ). Consider a set of groundtruth segments consisting of subset $S$ and subset $T$. $S$ and $T$ consist of segments corresponding to \textit{stuff} and \textit{thing} classes respectively. Similarly, we have predictions with subsets $S'$ and $T'$. For a given \textit{stuff} class $c$, we obtain the corresponding matching \textit{stuff} segments $MS_c=\{(s', s)\in S'_c\times S_c\,:\, \text{IoU}(s', s)>0\}$ as each image can have at most one predicted segment and at most one groundtruth segment. Thus, APQ\textsubscript{sc} corresponding to the \textit{stuff} class c is then computed as
\begin{equation}
\text{APQ}_{sc} = \frac{1}{|S_c|}\sum_{(s',s)\in MS_c}\text{IoU}(s',s),
\end{equation}
where $|S_c|$ is the total number of \textit{stuff} groundtruth segments corresponding to class $c$. The computed APQ\textsubscript{sc} follows the scheme suggested in~\cite{porzi2019seamless}.

Next, for a \textit{thing} class $c$ we obtain the matching segments by solving a maximum weighted bipartite matching problem~\cite{west2001introduction} for each pair  $(V, V')$ and $(O, O')$. Here,  $V$ and $O$ are the subsets of $T$ corresponding to the visible and occluded region segments. $V'$ and $O'$ are similar subsets of $T'$. This unique matching of segments splits the groundtruth and predicted \textit{thing} class segments ($T$ and $T'$) into three sets: matched pairs of segments (TP), unmatched groundtruth segments (FN), and unmatched predicted segments (FP). Hence APQ\textsubscript{tc} corresponding to the \textit{thing} class c is then defined as
\begin{equation}
\text{APQ}_{tc} = \frac{\sum_{(t',t)\in TP_c}\text{IoU}(t',t)}{|TP_c|+|FP_c|+|FN_c|}.
\end{equation}
Then, the overall APQ metric is the average over all the classes and is given by
\begin{equation}
\text{APQ} = \frac{\sum_{c\in C_s}APQ_{sc} + \sum_{c\in C_t}\text{APQ}_{tc}}{|C_s|+ |C_t|},
\end{equation}
where $C_s$ is the set of \textit{stuff} semantic classes and $C_t$ is the set of \textit{thing} semantic classes. Further, to explicitly analyze the performance of the model for visible and invisible or occluded regions, the $\text{APQ}_{tc}$ is comprised of $\text{APQ}_{vtc}$ and $\text{APQ}_{otc}$ which are computed with respect to the visible regions and occluded regions, respectively as
\begin{align}
\text{APQ}_{vtc} = \frac{\sum_{(v',v)\in TP_c}\text{IoU}(v',v)}{| TP_{cv}|+|FP_{cv}|+|FN_{cv}|},\\
\text{APQ}_{otc} = \frac{\sum_{(o',o)\in TP_c}\text{IoU}(o',o)}{| TP_{co}|+|FP_{co}|+|FN_{co}|},
\end{align}
where $v'$ and $o'$ are the visible and occluded regions of the predicted instance segments, $v$ and $o$ are the visible and occluded parts of the groundtruth instance segments. 


\subsection{Amodal Parsing Coverage}

The amodal panoptic quality metric is based on matching segments and as a consequence, it treats all the instances equally irrespective of their sizes. However, in some applications, a relatively higher segmentation quality of larger objects is more desirable than smaller objects such as in portrait segmentation and autonomous driving. This factor motivated Yang~\textit{et~al.}~\cite{yang2019deeperlab} to formulate the parsing covering (PC) metric for panoptic segmentation which accounts for the size of instances. We adapt the PC metric for amodal panoptic segmentation and propose the amodal parsing coverage (APC) metric. Let $P_c$ and $P_c'$ be the groundtruth and prediction for a $c$ semantic class respectively. If $c$ is a \textit{stuff} class, the coverage of \textit{stuff} class $c$ ($Cov_{sc}$) is computed similar to the coverage computation in PC, defined as
\begin{equation}
    Cov_{sc} = \frac{1}{N_c}\sum_{X\in P_{c}}\left | X \right | \cdot \max_{X'\in P_c'}IoU(X',X),
\end{equation}
where $N_c$ is the total number of pixels corresponding to class c in the groundtruth. For a \textit{thing} class $c$, the groundtruth segmentation $P_c$ and the predicted segmentation $P_c'$ are divided into visible segmentation $P_{vc}$ and invisible or occluded segmentation $P_{oc}$. Then the coverage ($Cov_{tc}$) for the \textit{thing} class $c$ is defined as
\begin{equation}
    Cov_{tc} = \frac{N_{vc} \cdot Cov_{vtc} + N_{oc} \cdot Cov_{otc}}{N_{vc}+N_{oc}},
\end{equation}
where $N_{vc}$ and $N_{oc}$ are the total numbers of pixels corresponding to class c in the groundtruth for visible and occluded regions respectively, and
\begin{align}
    Cov_{vtc} &=  \frac{1}{N_{vc}}\sum_{X\in P_{vc}}\left | X \right | \cdot \max_{X'\in P_{vc'}}IoU(X',X),\\
    Cov_{otc} &=  \frac{1}{N_{oc}}\sum_{X\in P_{oc}}\left | X \right | \cdot \max_{X'\in P_{oc'}}IoU(X',X).
\end{align}

Finally, APC is computed as the average over combined \textit{stuff} and \textit{thing} class coverage over all semantic classes as
\begin{equation}
\text{APC} = \frac{\sum_{c\in C_s}Cov_{sc} + \sum_{c\in C_t}Cov_{tc}}{|C_s|+ |C_t|},
\end{equation}
where $C_s$ and $C_t$ is the set of \textit{stuff} and  \textit{thing} semantic classes respectively. In summary, the proposed APC is devoid of any segment matching and incorporates the area weighted IoU to emphasize on the segmentation quality of large objects. Consequently, this metric accentuates segmentation quality of large occluded regions.

\section{Datasets}
In this section, we first give an overview of the annotation protocol that we employ for curating the amodal panoptic segmentation benchmark datasets followed by a brief description of each of the datasets. We choose the aforementioned datasets as they provide large-scale instance annotations that are consistent in time.



\subsection{Anotation Protocol}

We annotate two large-scale urban scene understanding datasets, KITTI-360 and BDD100K. We follow a semi-automatic annotation pipeline similar to~\cite{voigtlaender2019mots}. Specifically, we use the state-of-the-art EfficientPS~\cite{mohan2020efficientps} model pretrained on the Mapillary Vistas~\cite{neuhold2017mapillary} and Cityscapes~\cite{cordts2016cityscapes} datasets. We annotate images with pixel-level labels for amodal instance segmentation of \textit{thing} classes and semantic segmentation of \textit{stuff} classes. For amodal instance annotations, we fine-tune the pretrained EfficientPS model on the KINS dataset~\cite{qi2019amodal} which consists of amodal instance segmentation labels for urban road scenes. We generate pseudo amodal instance masks for a subset of the target dataset (BDD100K and KITTI-360). Subsequently, a human annotator manually corrects and refines these resulting pseudo labels. We then again fine-tune the EfficientPS model on the refined annotations and generate a new set of pseudo amodal instance masks for the next subset of the target dataset. We reiterate the aforementioned process until the entire dataset is fully annotated. Similarly, for semantic segmentation annotations, we fine-tune the pretrained EfficientPS model on the semantic segmentation labels of BDD100K. We then use this fine-tuned model to generate pseudo semantic segmentation labels of \textit{stuff} classes and follow the iterative semi-automatic annotation procedure. We adapt the publicly available labeling tool from~\cite{cordts2016cityscapes} for our manual annotations.

\subsection{KITTI-360-APS}

We extend the KITTI-360~\cite{Liao2021ARXIV} dataset which has semantic and instance labels with amodal panoptic annotations and name it the KITTI-360-APS dataset. It consists of nine sequences of urban street scenes with annotations for $61{,}168$ images of resolution $1408\times376$~pixels. Our dataset comprises $10$ \textit{stuff} classes. We define a class as \textit{stuff} if the class has amorphous regions or is incapable of movement at any point in time. Road, sidewalk, building, wall, fence, pole, traffic sign, vegetation, terrain, and sky are the stuff classes. Further, the dataset consists of $7$ \textit{thing} classes, namely car, pedestrians, cyclists, two-wheeler, van, truck, and other vehicles. Please note that we merge the bicycle and motorcycle class into a single class called two-wheelers. We use the sequence $10$ of the KITTI-360 dataset as the validation set and the rest of the sequences as the training set.

\begin{figure*}
    \centering
    \begin{subfigure}[b]{0.30\linewidth}
        \centering
        \includegraphics[width=\textwidth]{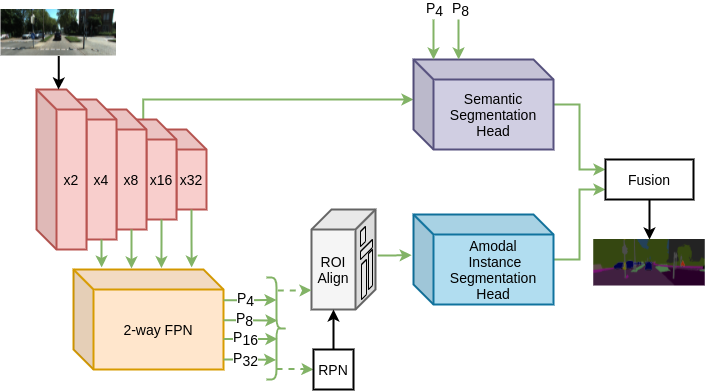}
        \subcaption{APSNet Architecture}
        \label{APSNEt_arch}
    \end{subfigure}
        \hfill
    \begin{subfigure}[b]{0.4\linewidth}
        \centering
        \includegraphics[width=\textwidth]{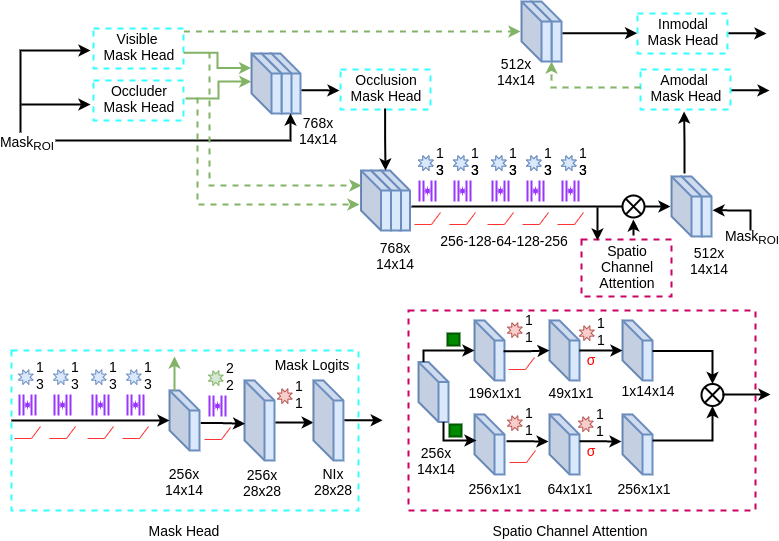}
        \subcaption{Amodal Instance Segmentation Head}
        \label{amodal_segmentation_head1}
    \end{subfigure}
    \hfill
    \begin{subfigure}[b]{0.25\linewidth}
        \centering
        \includegraphics[width=\textwidth]{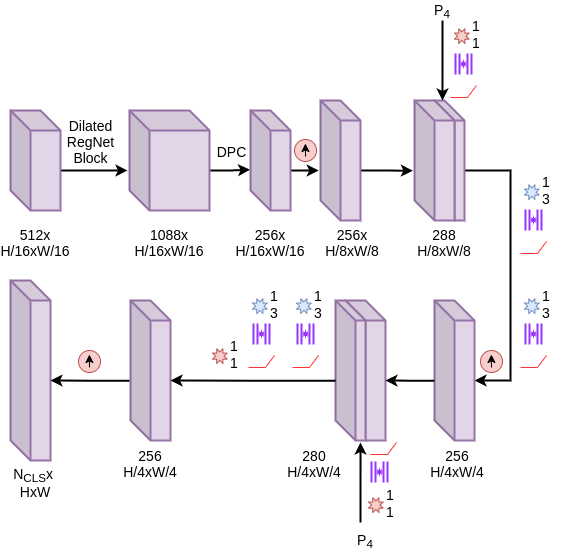}
        \subcaption{Semantic Segmentation Head}
        \label{semantic_segmentation_head}
    \end{subfigure}
    \\
    \vspace{0.2cm}
    \begin{subfigure}[b]{0.7\linewidth}
        \centering
        \includegraphics[width=\textwidth]{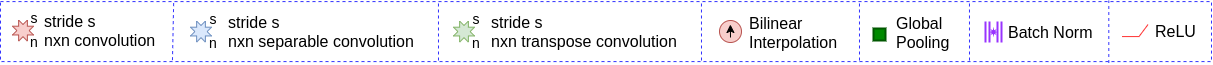}
        \label{panoptic_segmentation_eg1}
    \end{subfigure}
\vspace{-0.5cm}
    \caption{\subref{APSNEt_arch} Illustration of our proposed \mbox{APSNet} architecture consisting of a shared backbone and parallel semantic and amodal instance segmentation heads followed by a fusion module that fuses the outputs of both heads to yield the amodal panoptic segmentation output. \subref{semantic_segmentation_head} and \subref{amodal_segmentation_head1} present the topologies of architectural components of our proposed semantic segmentation head and amodal instance segmentation head respectively.}
    \label{fig:network}
    \vspace{-0.3cm}
\end{figure*}

\subsection{BDD100K-APS}

The Berkeley Deep Drive (BDD100K)~\cite{yu2018bdd100k} instance segmentation dataset comprises of $157$ training sequence and $39$ validation sequences. Each sequence contains $202$ images of resolution $1280\times720$ pixels with instance segmentation groundtruth labels. For our BDD100K-APS dataset, we select $12$ sequences from the training set and $3$ sequences from the validation set. We provide amodal panoptic annotations for $10$ \textit{stuff} classes and $6$ \textit{thing} classes. Road, sidewalk, building, fence, pole, traffic sign, fence, terrain, vegetation, and sky are the \textit{stuff} classes. Whereas, pedestrian, car, truck, rider, bicycle, and bus are the \textit{thing} classes.

\section{Baselines}
We introduce a total of six baselines for our proposed amodal panoptic segmentation task. We create the baselines by building upon the EfficientPS~\cite{mohan2020efficientps} model which is a state-of-the-art top-down panoptic segmentation network and replace its instance segmentation head with different existing amodal instance segmentation approaches. We choose the baseline's amodal head based on two aspects: the relevance of existing architectures to our task and the complexity involved in adapting the approach for our purpose. Hence, we adopt the following five state-of-the-art amodal instance segmentation methods for the instance head of our baselines: ORCNN~\cite{follmann2019learning}, VQ-VAE~\cite{jang2020learning}, Shape Prior~\cite{yuting2021amodal}, ASN~\cite{qi2019amodal}, and BCNet~\cite{Ke_2021_CVPR}. We introduce an additional baseline called Amodal-EfficientPS in which we add an extra amodal mask prediction layer to the instance head of the EfficientPS architecture. We use the post-processing step described in~\cite{mohan2020efficientps} to compute the panoptic segmentation output. We first obtain the amodal mask for each instance in the panoptic segmentation output using the amodal mask logits channels associated with the corresponding instance~ID. We then employ the sigmoid function on the selected amodal mask logits and threshold it at $0.5$ to obtain the final amodal binary mask. The set of the amodal binary mask along with its class prediction and instance ID is concatenated with the panoptic segmentation output to yield the final amodal panoptic prediction. We describe each of the architectures of the baselines and the post-processing step in detail in the supplementary material.

\section{APSNet Architecture}
In this section, we present a brief overview of our proposed \mbox{APSNet} architecture and then detail each of its constituting components. \figref{fig:network}~\subref{APSNEt_arch} depicts the topology of \mbox{APSNet} that follows the top-down approach. It consists of a shared backbone that comprises of an encoder and the 2-way Feature Pyramid Network (FPN)~\cite{mohan2020efficientps}, followed by the semantic segmentation head and amodal instance segmentation head. We employ the RegNet~\cite{radosavovic2020designing} architecture as the encoder (depicted in red). It consists of a standard residual bottleneck block with group convolutions. The overall architecture of this encoder consists of repeating units of the same block at a given stage and comprises a total of five stages. At the same time, it has fewer parameters in comparison to other encoders but with higher representational capacity. Subsequently, after the \mbox{2-way FPN}, our network splits into two parallel branches. One of the branches consists of the Region Proposal Network (RPN) and ROI align layers that take the \mbox{2-way FPN} output as input. The extracted ROI features after the ROI layers are propagated to the amodal instance segmentation head. The second parallel branch consists of the semantic segmentation head that is connected from the fourth stage of the encoder.

\subsection{Amodal Instance Segmentation Head}

Our proposed amodal instance segmentation comprises three parts, each focusing on one of the critical requirements for amodal reasoning. \figref{fig:network}~\subref{amodal_segmentation_head1} shows the architecture of our amodal instance segmentation head. First, the visible mask head learns to predict the visible region of the target object in a class-specific manner. Simultaneously, an occluder head, class-agnostically predicts the regions that occlude the target object. Specifically, the visible mask head learns to segment background objects for a given proposal and the occluder head learns to segment foreground objects. The occluder head provides a global initial guess estimate of where the occluded region of the target object exists. 

With the features from both visible and occluder mask heads, the amodal instance segmentation head can reason about the presence of the occluded region as well as its shape. This is achieved by employing an occlusion mask head that predicts the occluded region of the target object given the visible and occluder features. Specifically, the occlusion mask head takes the concatenated visible and occluder features along with $\text{Mask}_{\text{ROI}}$ features as input. We use the $\text{Mask}_{\text{ROI}}$ features as part of the input to the occlusion mask head to enable reasoning about the given proposal as a whole and not individual visible and occluder regions. Additionally, the occlusion mask head learns to predict the occluded region of the target object in a spatially independent manner. This allows the head to focus only on learning the underlying general shape relationship for a given visible and occluder region that completes the visible region to attain amodal perception. By focusing on the \textit{what} aspect of the occluded region (what should be the segmentation mask of the occluded region), we ease the learning of the occlusion mask head. Our method allows this ease in training due to denser feedback in contrast to sparser feedback in partial occlusion cases and hence enables capturing of the underlying shape of the occluded region effectively. We present the spatially dependent and independent groundtruth examples in the supplementary material.

Subsequently, the concatenated visible, occluder, and occlusion mask head features are further processed by a series of convolutions followed by a spatio-channel attention block. The spatio-channel attention block consists of two parallel branches. In one of the parallel branches, global pooling is applied spatially, we refer to this as the channel attention branch. The channel attention branch further consists of two $1\times1$ convolutions with $64$ and $256$ output channels respectively. The first $1\times1$ convolution has a ReLU activation and the second convolution has a sigmoid activation. The output of the channel attention branch is then multiplied with the output of the other parallel branch called the spatial branch. The spatial branch consists of a channel-wise global pooling layer, followed by reshaping the tensor from $1\times 14\times 14$ to $196\times1\times1$. Subsequently, two $1\times1$ convolutions are employed with $49$ and $196$ output channels respectively. The output is then reshaped to a $1\times 14\times 14$ tensor. Lastly, the output of the two branches is multiplied to compute the final output of the spatio-channel attention block. The aforementioned network layers aim to model the inherent relationship between the visible, occluder and occlusion features. Subsequently, these features are concatenated with the $\text{Mask}_{\text{ROI}}$ features to act as input to the amodal mask head. This amodal mask head then predicts the final amodal mask for the target object. Additionally, the visible mask is further refined using a second visible mask head that takes the concatenated amodal features and visible features to predict the final inmodal mask.

Lastly, our amodal instance segmentation head employs the Mask R-CNN bounding box head with two output heads: object classification and amodal bounding box. We use the binary cross-entropy loss for training each of the mask heads in our amodal instance segmentation head. The loss functions are described in detail in the supplementary material.

\subsection{Semantic Segmentation Head}

The architecture of our semantic segmentation head is illustrated in \figref{fig:network}~\subref{semantic_segmentation_head}. The semantic head takes the $\times16$ downsampled feature maps from the stage $4$ of the RegNet encoder as input. We employ an identical stage $5$ RegNet block with the dilation factor of the $3\times3$ convolutions set to $2$. We refer to this block as the dilated RegNet block. Subsequently, we employ a DPC~\cite{chen2018searching} module to process the output of the dilated block. We then upsample the output to $\times8$ and $\times4$ downsampled factor using bilinear interpolation. After each upsampling stage, we concatenate the output with the corresponding features from the \mbox{2-way FPN} having the same resolution and employ two $3\times3$ depth-wise separable convolutions to fuse the concatenated features. Finally, we use a $1\times1$ convolution to reduce the number of output channels to the number of semantic classes followed by a bilinear interpolation to upsample the output to the input image resolution. We employ the weighted per-pixel log-loss~\cite{bulo2017loss} for training similar to~\cite{mohan2020efficientps}.

\section{Experimental Evaluation}
In this section, we describe the training protocol that we use for the baselines and our proposed \mbox{APSNet} architecture. We then present extensive benchmarking results on KITTI-360-APS and BDD100K-APS in~\secref{sec:benchmark}. Subsequently, we present a detailed ablation study on the proposed amodal instance head in \secref{sec:aih}, followed by results for amodal instance segmentation on the KINS~\cite{qi2019amodal} dataset in \secref{sec:kins}. Finally, we present qualitative comparisons in \secref{sec:qualitative}.

\begin{table*}
\footnotesize 
\centering
\begin{tabular}{p{2.8cm}|p{0.4cm}p{0.4cm}p{0.5cm}p{0.5cm}p{0.5cm}p{0.5cm}p{0.3cm}p{0.6cm}|p{0.4cm}p{0.4cm}p{0.5cm}p{0.5cm}p{0.5cm}p{0.5cm}p{0.3cm}p{0.5cm}}
\toprule
Model &  \multicolumn{8}{c|}{KITTI-360-APS} & \multicolumn{8}{c}{BDD100K-APS}\\
\cmidrule{2-17}
 &  APQ  & APC & APQ$_S$ &APQ$_T$ & APC$_S$&  APC$_T$ & AP & mIOU & APQ  & APC & APQ$_S$ &APQ$_T$ & APC$_S$&  APC$_T$ & AP & mIOU\\
\midrule
Amodal-EfficientPS & $41.1$ & $57.6$ & $46.2$ & $33.1$ & $58.1$ & $56.6$ & $29.1$ & $44.7$ & $44.9$ & $46.2$ & $54.9$ & $29.9$ & $64.7$ & $41.4$ & $25.6$ & $50.4$ \\
ORCNN~\cite{follmann2019learning}  & $41.1$ & $57.5$ & $46.2$ & $33.1$ & $58.1$ & $56.6$ & $29.0$ & $44.5$ & $44.9$ & $46.2$ & $54.9$ & $29.9$ & $64.7$ & $41.5$ & $25.6$ & $50.4$ \\
BCNet~\cite{Ke_2021_CVPR}  & $41.6$ & $57.9$ & $46.2$ & $34.4$ & $58.1$ & $57.6$ & $30.3$ & $45.8$ & $45.2$ & $46.4$ & $55.0$ & $30.7$ & $64.7$ & $42.1$ & $26.3$ & $51.0$ \\
VQ-VAE~\cite{jang2020learning}  & $41.7$ & $58.0$  & $46.2$ & $34.6$ & $58.1$ & $57.8$ & $30.4$ & $45.9$ & $45.3$ & $46.5$ & $54.9$ & $30.8$ & $64.7$ & $42.2$ & $27.3$ & $51.1$  \\
Shape Prior~\cite{yuting2021amodal}  & $41.8$ & $58.2$  & $46.2$ & $35.0$ & $58.1$ & $58.2$ & $31.0$ & $46.3$ & $45.4$ & $46.6$ & $55.0$ & $31.0$ & $64.8$ & $42.6$ & $27.6$ & $52.4$  \\
ASN~\cite{qi2019amodal} & $41.9$ & $58.2$ & $46.2$ & $35.2$ & $58.1$ & $58.3$ & $31.1$ & $46.3$ & $45.5$ & $46.6$ & $55.0$ & $31.2$ & $64.8$ & $42.7$ & $27.9$ & $52.5$   \\
\midrule
\mbox{APSNet} (Ours)  &  $\mathbf{42.9}$ & $\mathbf{59.0}$  & $\mathbf{46.7}$ & $\mathbf{36.9}$  & $\mathbf{58.5}$ & $\mathbf{59.9}$  & $\mathbf{33.4}$ &$\mathbf{48.0}$ & $\mathbf{46.3}$ & $\mathbf{47.3}$  & $\mathbf{55.4}$ & $\mathbf{32.8}$  & $\mathbf{65.1}$ & $\mathbf{44.5}$  & $\mathbf{29.2}$ & $\mathbf{53.3}$ \\
\bottomrule
\end{tabular}
\vspace{-0.2cm}
\caption{Performance comparison of amodal panoptic segmentation on the KITTI-360-APS and BDD100K-APS validation set. Subscripts $S$ and $T$ refer to \textit{stuff} and \textit{thing} classes respectively. All scores are in [\%].}
\label{tab:kittiEvaluation}
\vspace{-0.3cm}
\end{table*}

We use PyTorch~\cite{paszke2019pytorch} for implementing all our architectures and we trained our models on a system with an Intel Xenon (2.20GHz) processor and NVIDIA TITAN RTX GPUs. We train our network on two crop resolutions of the input image according to the dataset. We use crops of $376\times1408$~pixels and $448\times1280$~pixels for the KITTI-360-APS and BDD100K-APS dataset respectively. We use a multi-step learning rate schedule with a drop factor of 10. We use a base learning rate of $0.04$ and $0.01$ for KITTI-360-APS and BDD100k-APS respectively. We train our model on the KITTI-360-APS dataset for $40$ epochs and $200$ epochs on the BDD100K-APS dataset. We set the milestones as $65\%$ and $90\%$ of the total epochs.

\subsection{Benchmarking Results}
\label{sec:benchmark}

In this section, we report results comparing the performance of our proposed \mbox{APSNet} architecture against the introduced baselines. For comparisons on KITTI-360-APS and BDD100K-APS, we report results of the models that we trained using the official implementations that have been publicly released by the authors and performed extensive tuning of hyperparameters to the best of our ability. We report results on the validation sets for all the datasets. \tabref{tab:kittiEvaluation} presents the benchmarking results.

In the baselines, all the other components of the amodal panoptic segmentation network remain the same except for the amodal instance head. Therefore, all the baselines achieve the same APQ$_S$ and APC$_S$ scores. In contrast, our \mbox{APSNet} model that incorporates our proposed semantic head achieves higher APQ$_S$ and APC$_S$ scores. This gain of $0.3\%\text{-}0.5\%$ in the aforementioned metrics demonstrate the better \textit{stuff} segmentation performance of our architecture. The improvement can be attributed to the ability of our semantic head to increase the receptive field for effective context aggregation by increasing the dilation factor of the subsequent encoding block that outputs features corresponding to $\times16$ downsampling factor instead of further downsampling. As a consequence, our network does not lose the ability to localize features, providing the decoder with better semantic features to use during the upsampling stage.\looseness=-1

Among the baselines, the ASN model achieves the highest APQ and APC scores. This method focuses on incorporating the global occlusion context in the model-specific mask prediction heads. The other baselines either capture occlusion features implicitly or learn the occlusion map but do not use the information in the mask prediction heads. The performance of the ASN model demonstrates the importance of incorporating explicitly modeled occlusion features for improved amodal reasoning. Nevertheless, our \mbox{APSNet} outperforms ASN in all the metrics, namely APQ and APC along with the sub-components of the metrics on both datasets. Moreover, it also achieves the highest AP and mIoU scores. These improvements can be partially attributed to the semantic head but the majority of the contribution is due to the proposed amodal instance head. The explicit coarse modeling of occlusion regions with occluder features and the spatially independent modeling of the occluded region given the visible and occluder features provides our amodal mask prediction head with additional cues that positively supplement its amodal reasoning abilities. Hence, our proposed \mbox{APSNet} architecture achieves state-of-the-art performance for the task of amodal panoptic segmentation.

We further analyze the relationship between the different metrics reported. Although the metrics assess different aspects of amodal scene parsing, due to the close relationship of these aspects, the metrics are positively correlated. This is evident from the reported results. With the increase in the APQ score, the APC score is likely to increase and vice-versa. This relationship also extends to the AP and mIoU metrics. Additionally, computing both metrics can be beneficial as the gain or loss proportion in each of the metrics provides more insights. APQ evaluates the amodal parsing quality independent of instance sizes whereas APC emphasizes segmentation quality of larger area instances. Thus, a higher gain in APQ compared to APC can indicate that the amodal segmentation quality of smaller object instances improves greatly compared to larger objects and vice-versa. We further explain this observation with the visible and occluded components of the metrics in the supplementary material. 



\begin{figure*}
\centering
\footnotesize
{\renewcommand{\arraystretch}{0.5}
\begin{tabular}{p{0.4cm}P{4.6cm}P{4.6cm}P{4.6cm}}
&  \raisebox{-0.4\height}{ASN~\cite{qi2019amodal}} &  \raisebox{-0.4\height}{\mbox{APSNet} (Ours)} & 
\raisebox{-0.4\height}{Improvement\textbackslash{Error Map}}\\
\\
\rot{(a)} 
& \raisebox{-0.4\height}{\includegraphics[width=\linewidth]{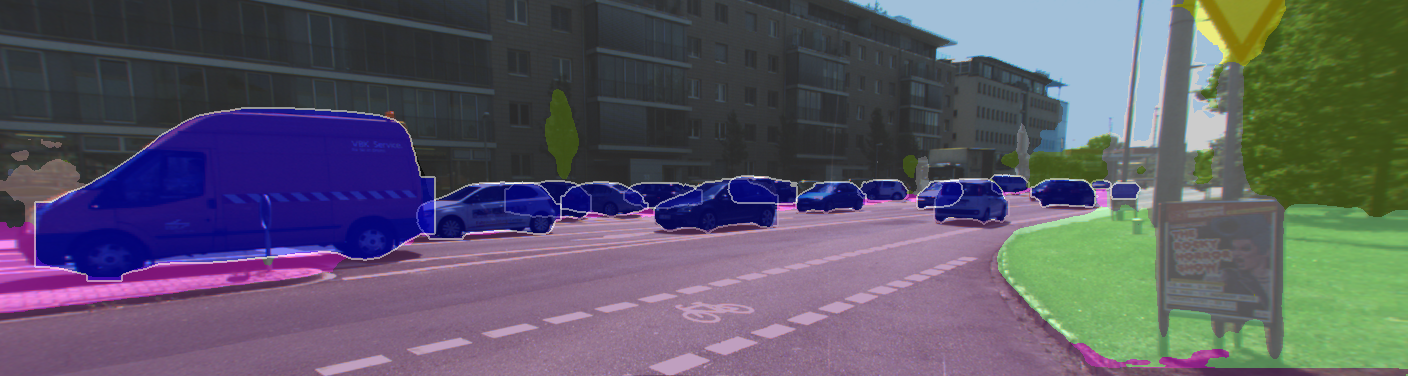}} & \raisebox{-0.4\height}{\includegraphics[width=\linewidth]{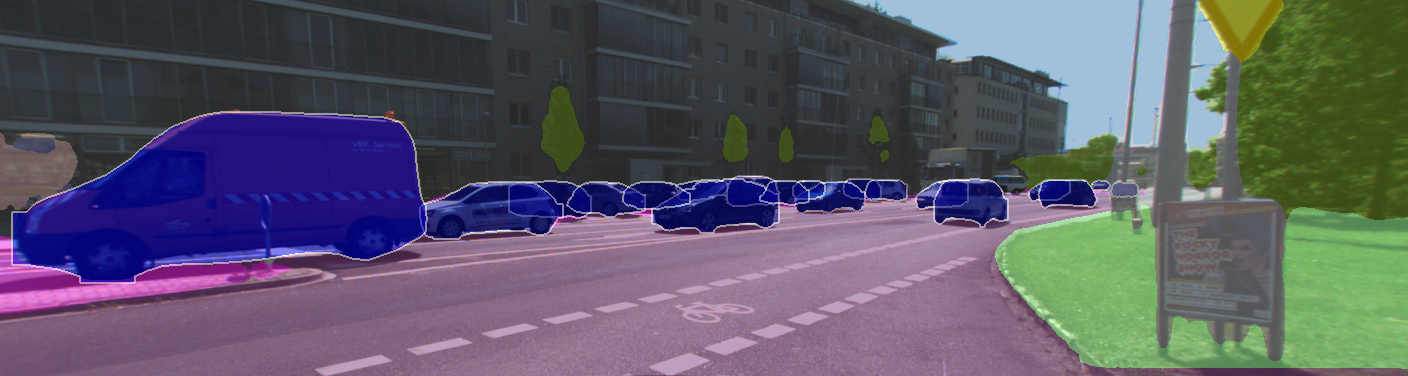}} & \raisebox{-0.4\height}{\includegraphics[width=\linewidth,frame]{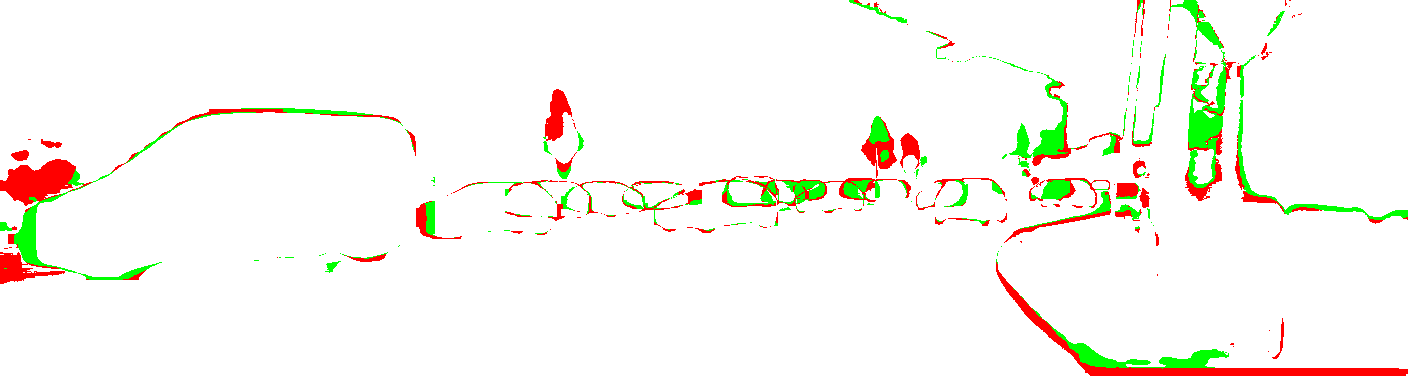}}\\
\\
\rot{(b)} 
& \raisebox{-0.4\height}{\includegraphics[width=\linewidth]{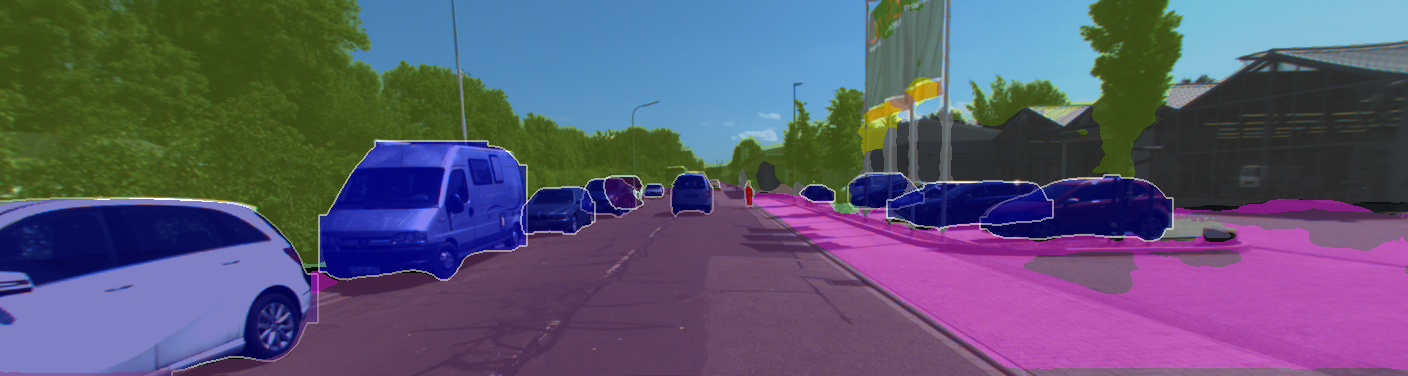}} & \raisebox{-0.4\height}{\includegraphics[width=\linewidth]{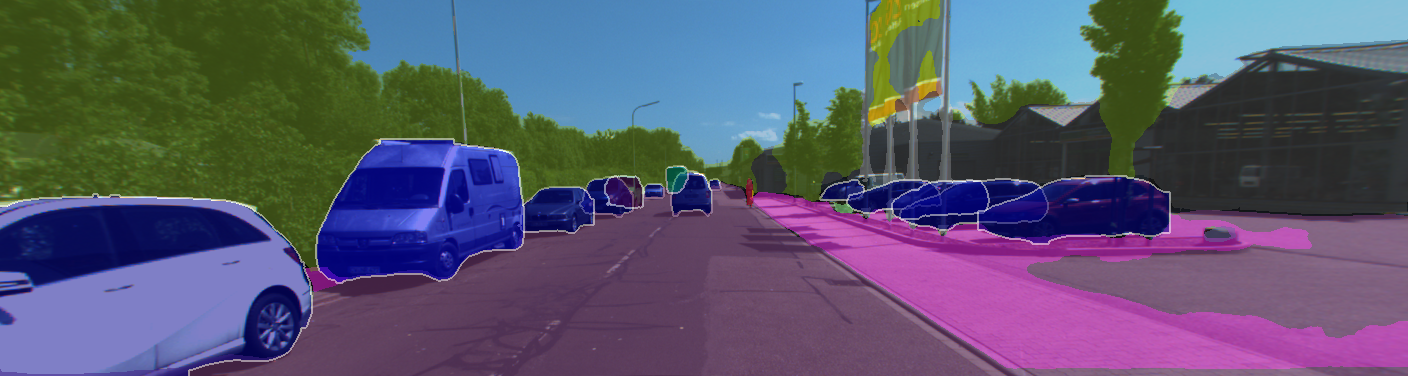}} & \raisebox{-0.4\height}{\includegraphics[width=\linewidth,frame]{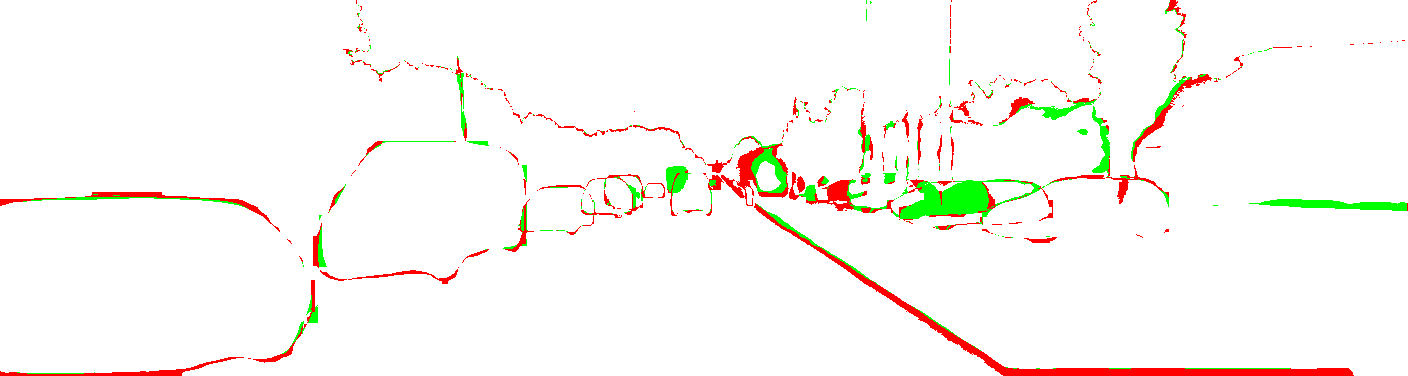}}\\
\\
\rot{(c)} 
& \raisebox{-0.4\height}{\includegraphics[width=\linewidth]{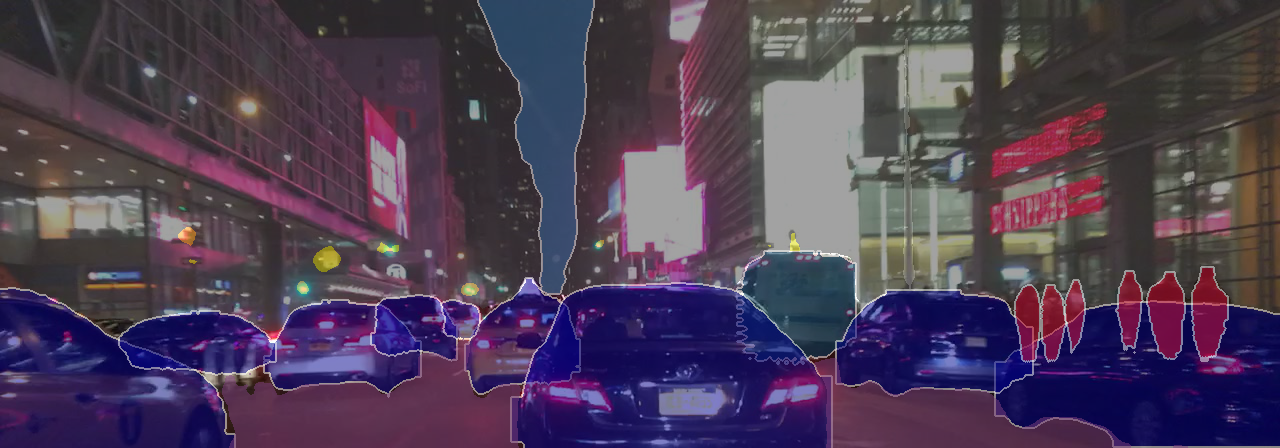}} & \raisebox{-0.4\height}{\includegraphics[width=\linewidth]{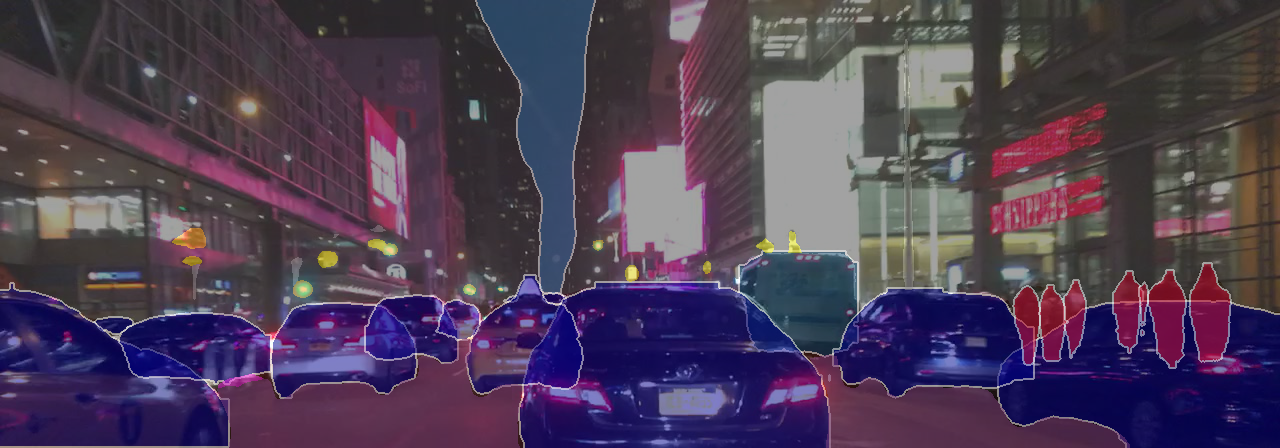}} & \raisebox{-0.4\height}{\includegraphics[width=\linewidth,frame]{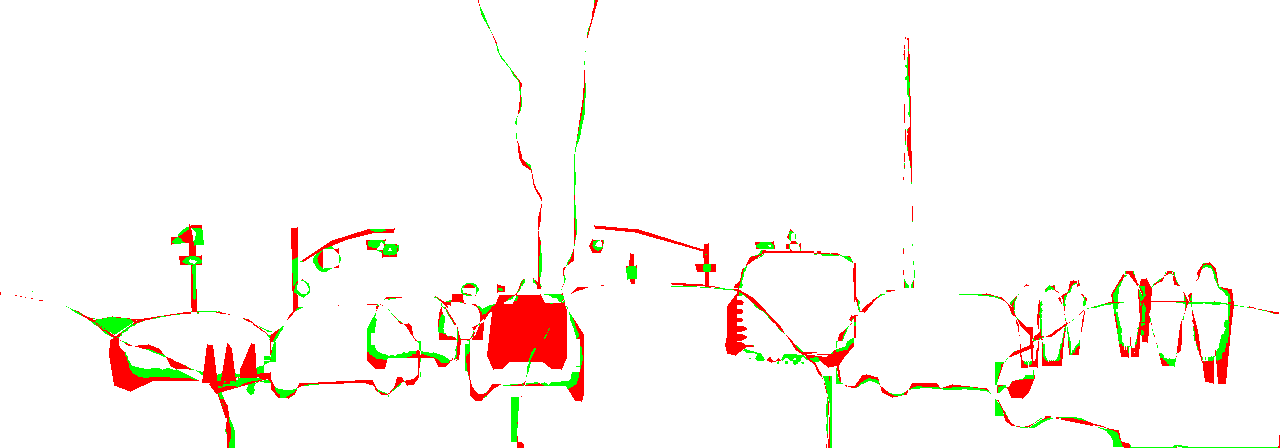}}\\
\\
\rot{(d)} 
& \raisebox{-0.4\height}{\includegraphics[width=\linewidth]{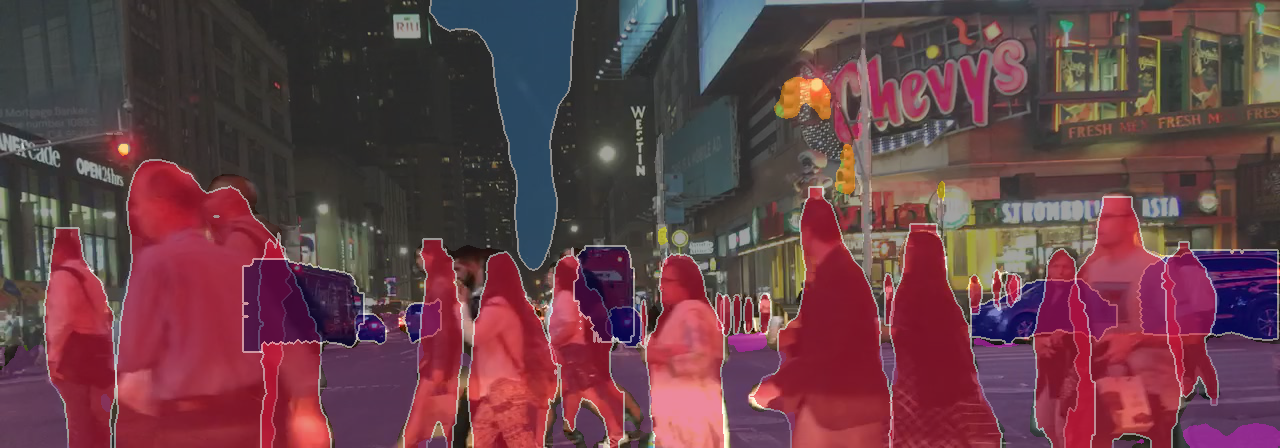}} & \raisebox{-0.4\height}{\includegraphics[width=\linewidth]{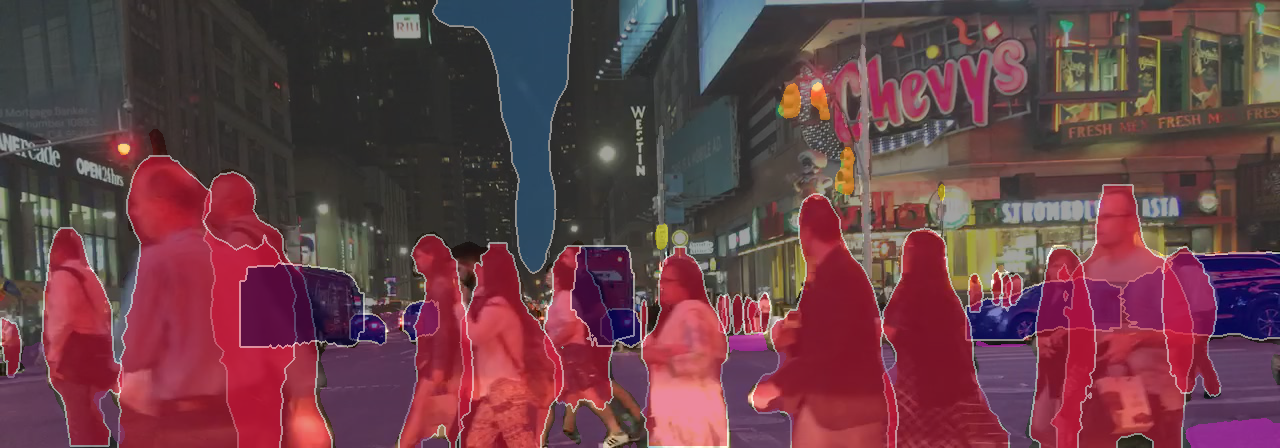}} & \raisebox{-0.4\height}{\includegraphics[width=\linewidth,frame]{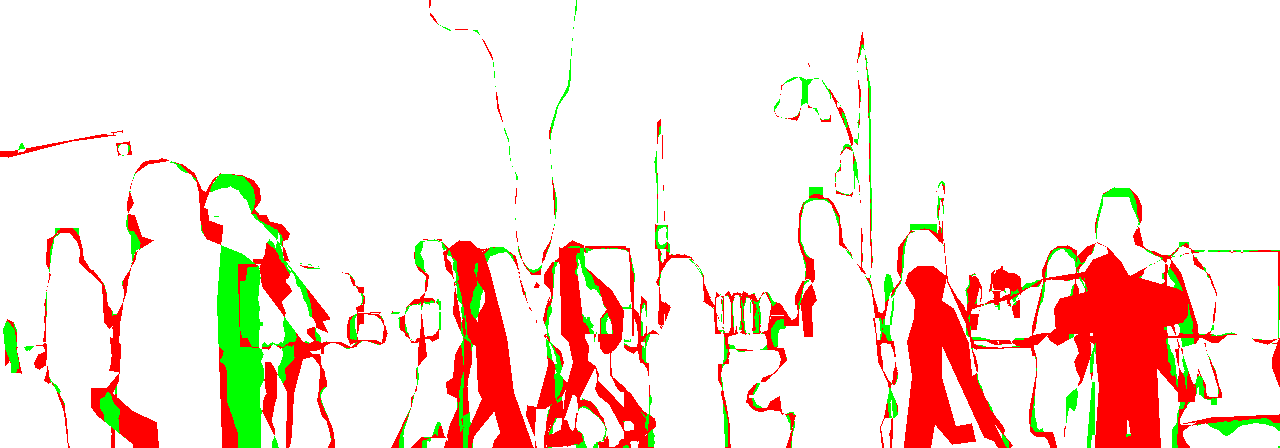}}\\
\\
\end{tabular}}
\vspace{-0.3cm}
\caption{Qualitative amodal panoptic segmentation results of our proposed \mbox{APSNet} network in comparison to the state-of-the-art baseline ASN~\cite{qi2019amodal} on KITTI-360-APS (a, b) and BDD100K-APS (c, d) datasets. We also show Improvement\textbackslash{Error Map} which denotes the pixels that are misclassified by \mbox{APSNet} in red and the pixels that are misclassified by the baseline but correctly predicted by \mbox{APSNet} in green.} 
\label{fig:visual_ablation}
\vspace{-0.2cm}
\end{figure*}

\subsection{Ablation Study on Amodal Instance Head}
\label{sec:aih}

\begin{table}
\footnotesize 
\centering
\begin{tabular}{p{2.0cm}|p{0.5cm}p{0.5cm}p{0.5cm}p{0.5cm}p{0.5cm}p{0.5cm}}
\toprule
Model &  APQ$_T$&  APQ$^V_T$ &  APQ$^O_T$ & APC$_T$& APQ$^V_T$ & APC$^O_T$ \\
\midrule
M1  &$33.3$ & $41.3$ &$15.1$ &  $56.9$ &$59.3$ & $23.4$   \\
M2  &$33.7$ & $41.4$ &$15.4$ &  $57.5$ &$59.4$ & $23.9$   \\
M3  & $34.6$ & $41.7$ &$15.7$ &  $58.2$ &$59.6$ & $24.4$  \\
M4  &$35.0$ & $42.6$ &$15.7$ &  $58.8$ &$60.2$ & $24.5$   \\
M5  &$35.9$ & $43.6$ &$17.7$ &  $59.4$ &$61.6$ & $25.1$   \\
M6 (Ours)  &$\mathbf{36.9}$ & $\mathbf{44.1}$ &$\mathbf{18.6}$ &  $\mathbf{59.9}$ &$\mathbf{62.2}$ & $\mathbf{25.8}$   \\
\bottomrule
\end{tabular}
\vspace{-0.2cm}
\caption{Evaluation of various architectural components of our proposed amodal instance segmentation head. The performance is shown for the models trained on the KITTI-360 APS dataset and evaluated on the validation set. Subscript $T$ refers to \textit{thing} classes. Superscripts $V$ and $O$ refer to visible and occluded regions respectively. All scores are in [\%].}
\label{tab:instanceHeadEvaluation}
\vspace{-0.4cm}
\end{table}

In this section, we quantitatively demonstrate the importance of each component of our proposed amodal instance head. \tabref{tab:instanceHeadEvaluation} presents results from this experiment. We report the metric's \textit{thing} component and its sub-components. We begin with the model M1 that employs visible or inmodal, and amodal mask prediction heads in the amodal instance head. In the M2 model, we then employ occlusion and visible mask prediction heads on top of which we add an amodal mask prediction head. The improvement in performance shows that modeling visible and occlusion features explicitly improves the amodal reasoning ability. Subsequently, in model M3, we add an occluder mask prediction head in parallel to the occlusion and visible mask prediction head of M2. The amodal mask prediction head is now built on top of these three mask prediction heads. The larger increase in the APC$^O_T$ score demonstrates that the occlusion region segmentation of nearby objects greatly improves the performance compared to faraway objects. The occluder features that are incorporated enable the amodal mask head to discern the boundaries of the occluded regions. In the M4 model, we add another visible mask prediction head that builds upon the visible and amodal mask heads. M4 achieves an improvement in APQ$^V_T$ by $0.9\%$ and in APC$^V_T$ by $0.6\%$. Building upon M4, in the M5 model, we predict spatially independent occlusion masks in addition to a processing block before the amodal mask head. Lastly, in the M6 model, following the processing block, we add the spatio-channel attention block. The improvement in results demonstrates that the processing block generates salient features for the amodal masks which are further enhanced by explicitly modeling the interdependencies between the channels and the spatial correlations of its features.

\subsection{Performance on KINS dataset}\label{sec:kins}
\begin{table}
\footnotesize 
\centering
\begin{tabular}{p{2.5cm}|p{1.2cm}p{1.2cm}}
\toprule
Model &  Amodal$_{AP}$ & Inmodal$_{AP}$ \\
\midrule
ORCNN~\cite{follmann2019learning}  & $29.0$ &$26.4$  \\
VQ-VAE~\cite{jang2020learning}  & $31.5$ &$-$  \\
Shape Prior~\cite{yuting2021amodal}  & $32.1$ &$29.8$  \\
ASN~\cite{qi2019amodal}  & $32.2$ &$29.7$   \\
\midrule
\mbox{APSNet} (Ours)  & $\mathbf{35.6}$  & $\mathbf{32.7}$ \\
\bottomrule
\end{tabular}
\vspace{-0.2cm}
\caption{Amodal instance segmentation results on the KINS dataset. All scores are in [\%].}
\label{tab:kins}
\vspace{-0.4cm}
\end{table}

KINS~\cite{qi2019amodal} is a benchmark for amodal instance segmentation. We evaluate the performance of \mbox{APSNet} on this sub-task of the proposed amodal panoptic segmentation by discarding its semantic segmentation head. This benchmark uses the AP metric for evaluating both amodal and inmodal segmentation. \tabref{tab:kins} presents results in which we observe that \mbox{APSNet} outperforms the state-of-the-art by $3.4\%$ and $2.9\%$ for amodal and inmodal AP respectively. This demonstrates that our proposed amodal instance head in \mbox{APSNet} also improves the inmodal segmentation performance.

\subsection{Qualitative Evaluations}
\label{sec:qualitative}

In this section, we qualitatively compare the amodal panoptic segmentation performance of our proposed \mbox{APSNet} with the best performing baseline ASN. \figref{fig:visual_ablation} presents the qualitative results. We observe that both approaches are capable of segmenting partial occlusion cases. However, our \mbox{APSNet} outperforms ASN under partial to moderate occlusion cases such as cluttered cars and pedestrians. Moreover, \mbox{APSNet} achieves better boundary segmentation of visible regions due to the refinement stage of the inmodal mask. The results of our proposed architecture are highly motivating, however the segmentation quality near the boundaries of moderately to heavily occluded regions of non-rigid classes such as pedestrians tends to be poor. These cases are extremely hard to predict for humans as well. However, humans can predict the occluded region with a high degree of consistency~\cite{zhu2017semantic}. We hope that this work encourages innovative solutions in the future to address this problem as well as other challenges of amodal panoptic segmentation.

\section{Conclusion}
In this work, we introduced and addressed the task of amodal panoptic segmentation. We formulated two easily interpretable evaluation metrics for measuring the performance of our proposed task. We introduced several strong baselines for amodal panoptic segmentation by combining state-of-the-art individual models of the sub-tasks. Further, we proposed the novel \mbox{APSNet} architecture that achieves state-of-the-art performance for amodal panoptic segmentation and amodal instance segmentation. We believe that these results demonstrate the feasibility of this ultimate scene parsing task and encourage new research avenues in the future.





{\small
\bibliographystyle{ieee_fullname}
\bibliography{egbib}
}

\flushcolsend
\pagebreak

\begin{strip}
\begin{center}

\vspace{-5ex}
\textbf{\Large \bf
Amodal Panoptic Segmentation} \\
\vspace*{12pt}

\Large{\bf Supplementary Material}\\
\vspace*{12pt}
\large{Rohit Mohan \qquad Abhinav Valada}\\
\large{University of Freiburg}\\
\vspace*{2pt}
\tt\small{\{mohan, valada\}@cs.uni-freiburg.de}

\end{center}
\end{strip}

\setcounter{section}{0}
\setcounter{equation}{0}
\setcounter{figure}{0}
\setcounter{table}{0}
\makeatletter

%


\normalsize

In this supplementary material, we provide additional details on various aspects of our work. We present dataset statistics for our proposed amodal panoptic segmentation datasets in \secref{sec:supp_dataset}. We then discuss the baseline architectures and the inference in-depth in \secref{sec:supp_baseline} and \secref{sec:supp_inference}, respectively. Subsequently, we provide details on the loss functions that we employ to train the amodal instance segmentation head of our APSNet in \secref{sec:supp_amodal}. Finally, we discuss the benchmarking results on the KITTI-360-APS dataset in detail to reinforce the utility of our proposed evaluation metrics in \secref{sec:supp_benchmark}.

\section{Datasets}
\label{sec:supp_dataset}

In this section, we present statistics and examples for each of the datasets that we introduce. To evaluate the shape complexity of the amodal segments, we compute the shape convexity and simplicity~\cite{zhu2017semantic} for each dataset as follows:
\begin{align}
  convexity(S) &= \frac{Area(S))}{Area(ConvexHull(S))}, \\
  simplicity(S) &= \frac{\sqrt{4\pi*Area(S)}}{Perimeter(S)}.
\end{align}

\tabref{tab:shape} presents the shape complexity metric scores for KITTI-360-APS and BDD-APS datasets. Additionally, we compare the convexity and simplicity of our dataset with existing amodal instance segmentation datasets namely COCO-A~\cite{zhu2017semantic} and KINS~\cite{qi2019amodal}.

\subsection{KITTI-360-APS}

\begin{figure}
    \centering
    \begin{subfigure}[b]{\linewidth}
        \centering
        \includegraphics[width=\textwidth]{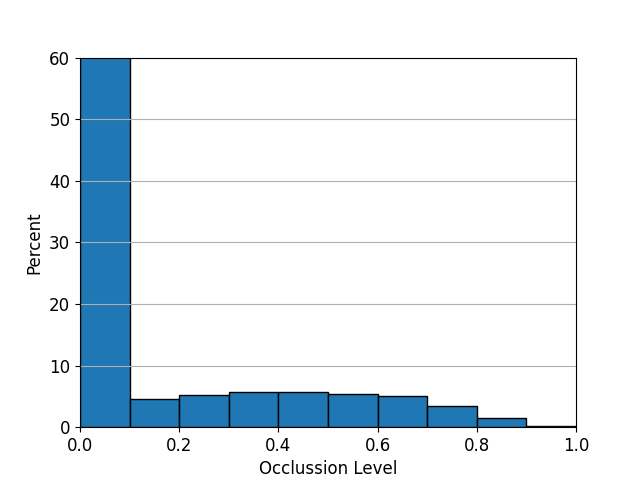}
        \subcaption{Occlusion level of KITTI-360-APS dataset.}
        \label{occ_kitti}
    \end{subfigure}
    \begin{subfigure}[b]{\linewidth}
        \centering 
        \includegraphics[width=\textwidth]{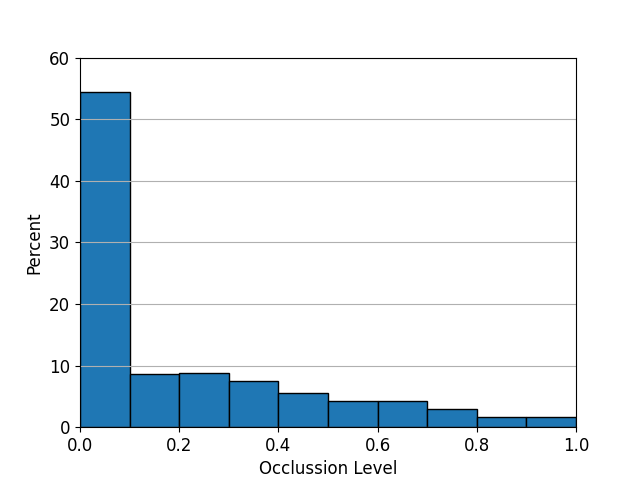}
        \subcaption{Occlusion level of BDD100K-APS dataset.}
        \label{occ_bdd}
    \end{subfigure}
    \caption{Illustration of occlusion level (defined as the fraction of region area that is occluded) in KITTI-360-APS~\subref{occ_kitti} and BDD100K-APS~\subref{occ_bdd} datasets. }
    \label{fig:occ}
\end{figure}

The KITTI-360-APS dataset consists of $11$ \textit{stuff} classes namely road, sidewalk, building, wall, fence, pole, traffic sign, vegetation, terrain, and sky. The dataset further comprises $7$ \textit{thing} classes, namely car, pedestrians, cyclists, two-wheeler, van, truck, and other vehicles. \tabref{tab:thingdist} presents the \textit{thing} class distribution for the dataset. We observe that the instances of the car are predominant in \textit{thing} classes followed by pedestrian and truck classes. The contribution of the Other-Vehicle class to the number of instances is the least with $0.2\%$. \figref{fig:occ}~\subref{occ_kitti} illustrates the histogram of occlusion level which is defined as the fraction of occluded region area. We notice about $60\%$ of the instances are either slightly occluded or not occluded at all in the dataset and the rest of the instances have different degrees of occlusions. The second peak in the graph is observed for near moderate occlusion levels while heavily occluded regions are relatively small in comparison. In terms of shape complexity (\tabref{tab:shape}), KITTI-360-APS consists of relatively simpler amodal segments indicated by the higher the convexity-simplicity average value which is in line with the intuition~\cite{zhu2017semantic} that independent of scene geometry and occlusion patterns, amodal segments tend to be relatively simpler. \figref{fig:visual_gt} presents examples from our dataset.

\begin{table}
\centering\small\renewcommand\arraystretch{1}
\renewcommand{\tabcolsep}{2mm}
\begin{tabular}{l|cc|cc}
\toprule
 & \multicolumn{2}{c|}{{Simplicity}} & \multicolumn{2}{c}{{Convexity}}\\
\cmidrule{2-5}
 & Inmodal & Amodal & Inmodal & Amodal \\
\midrule
COCO-A~\cite{zhu2017semantic} & 0.746   & 0.856   & 0.658   & 0.685\\
KINS~\cite{qi2019amodal}  & 0.709   & 0.830   & 0.610   & 0.639\\
KITTI-360-APS & 0.778   & 0.884   & 0.689   & 0.746\\
BDD100K-APS  & 0.697   & 0.821   & 0.594   & 0.618\\
\bottomrule
\end{tabular}
\caption{Comparison of shape statistics between inmodal and amodal segments in our proposed KITTI-360-APS and BDD100K datasets, along with COCO-A and KINS datasets.}
\label{tab:shape}
\end{table}

\begin{table*}
\footnotesize 
\centering
\begin{tabular}{l|ccccccc}
\toprule
Class & Car & Pedestrian & Cyclist & Two-Wheelers & Truck & Van & Other-Vehicles \\
\midrule
 Number & $192624$ & $6240$ &  $3096$ & $2805$ & $6561$ & $3573$ & $443$  \\
 Ratio & $89.4\%$ & $2.8\%$ &  $1.4\%$ & $1.3\%$ & $3.0\%$ & $1.6\%$ & $0.2\%$ \\
\bottomrule
\end{tabular}
\caption{\textit{Thing} class distribution of KITTI-360-APS dataset.}
\label{tab:thingdist}
\end{table*}

\begin{table*}
\footnotesize 
\centering
\begin{tabular}{l|cccccc}
\toprule
Class & Pedestrian & Car & Truck & Rider & Bicycle & Bus \\
\midrule
 Number & $19671$ & $23775$ &  $2653$ & $561$ & $1110$ & $1288$  \\
 Ratio & $40.1\%$ & $48.4\%$ &  $5.4\%$ & $1.1\%$ & $2.3\%$ & $2.7\%$  \\
\bottomrule
\end{tabular}
\caption{\textit{Thing} class distribution of BDD100K-APS dataset.}
\label{tab:thingdistbdd}
\end{table*}

\subsection{BDD100K-APS}

The BDD100K-APS dataset provides amodal panoptic annotations for $10$ \textit{stuff} classes and $6$ \textit{thing} classes. Road, sidewalk, building, fence, pole, traffic sign, fence, terrain, vegetation, and sky are the \textit{stuff} classes. Whereas, pedestrian, car, truck, rider, bicycle, and bus are the \textit{thing} classes. In the \mbox{BDD100K-APS} dataset, the number of instances of car and pedestrian classes is relatively close and are the predominant classes followed by the truck class.  Bicycle and bus classes have similar instance distributions whereas instances of rider are the least with $1.1\%$ of the total instances. \figref{fig:occ}~\subref{occ_bdd} presents the occlusion level distribution of instances of this dataset. About $54\%$ of the instances in the dataset are not occluded or are slightly occluded. The number of instances having a higher degree of occlusion level approximately decreases with an increase in the occlusion level. In \tabref{tab:thingdist}, the convexity-simplicity average value for the amodal segments is lower for this dataset implying BDD100K-APS is a more complex dataset due to the presence of a large number of non-rigid objects such as pedestrians.

\begin{figure*}
\centering
\footnotesize
{\renewcommand{\arraystretch}{0.5}
\begin{tabular}{P{8cm}P{8cm}}
 \raisebox{-0.4\height}{\includegraphics[width=\linewidth]{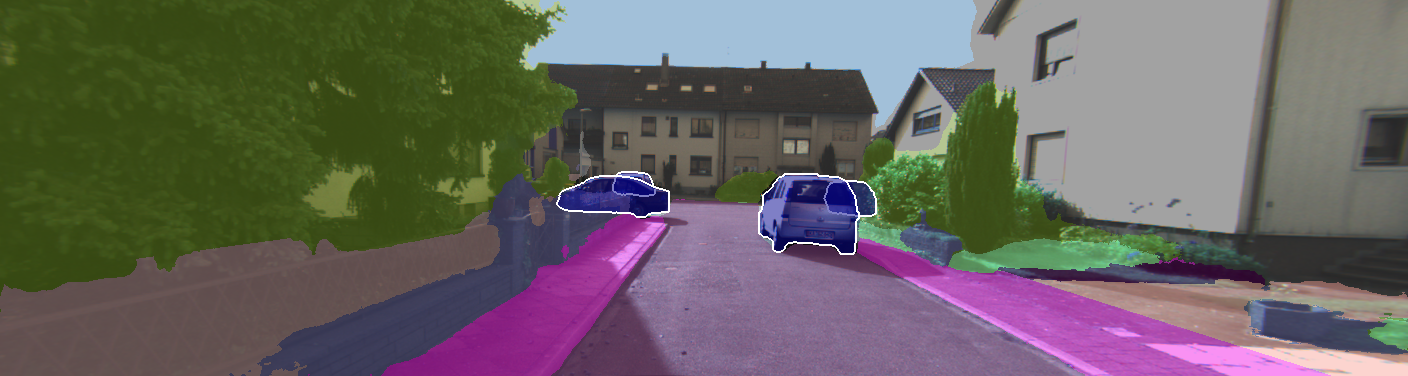}} & \raisebox{-0.4\height}{\includegraphics[width=\linewidth]{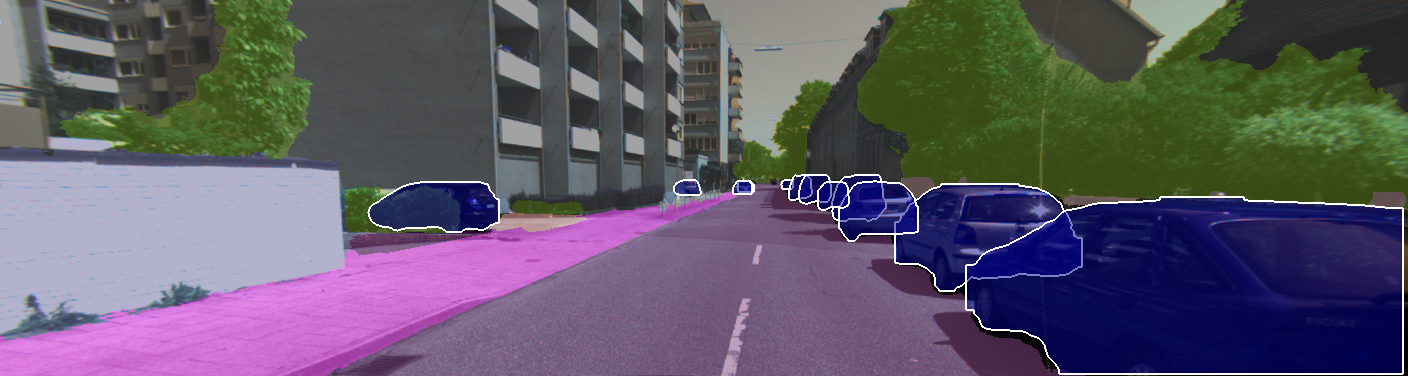}}\\ 
\\
(a) & (b) \\
\\
 \raisebox{-0.4\height}{\includegraphics[width=\linewidth]{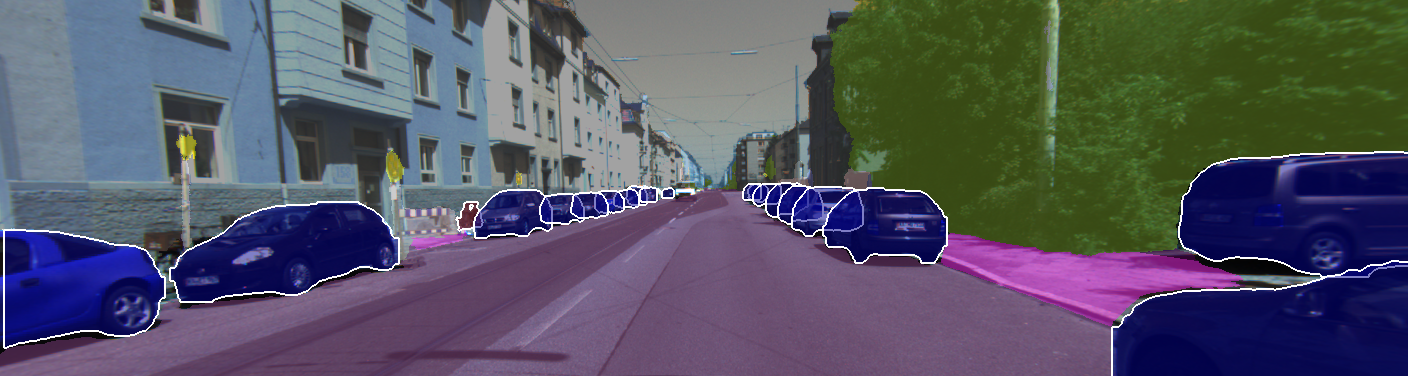}} & \raisebox{-0.4\height}{\includegraphics[width=\linewidth]{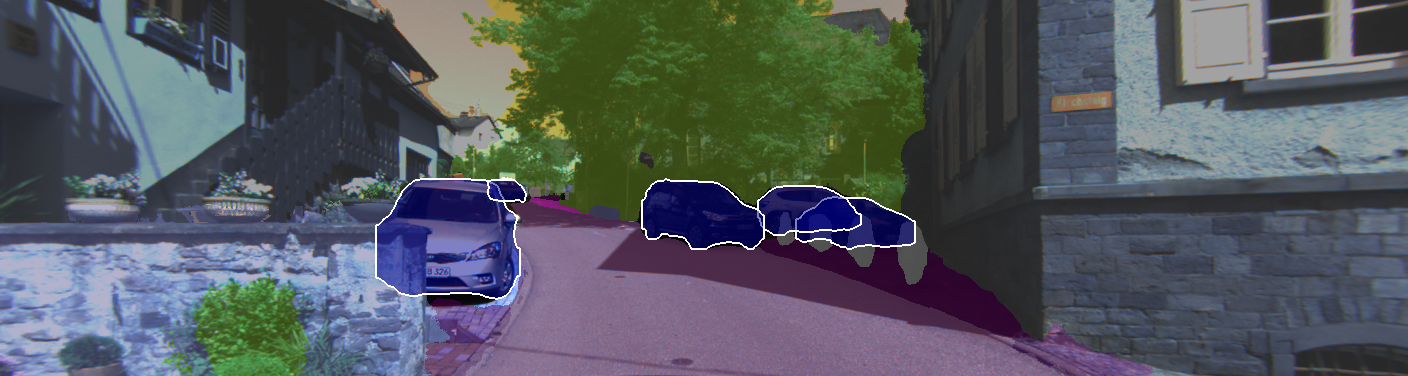}}\\ 
\\
(c) & (d) \\
\\
 \raisebox{-0.4\height}{\includegraphics[width=\linewidth]{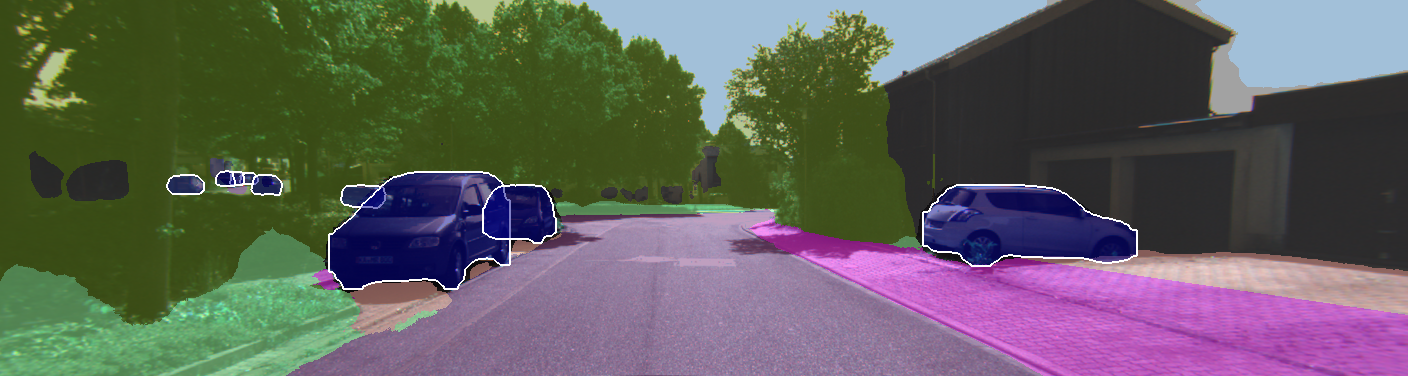}} & \raisebox{-0.4\height}{\includegraphics[width=\linewidth]{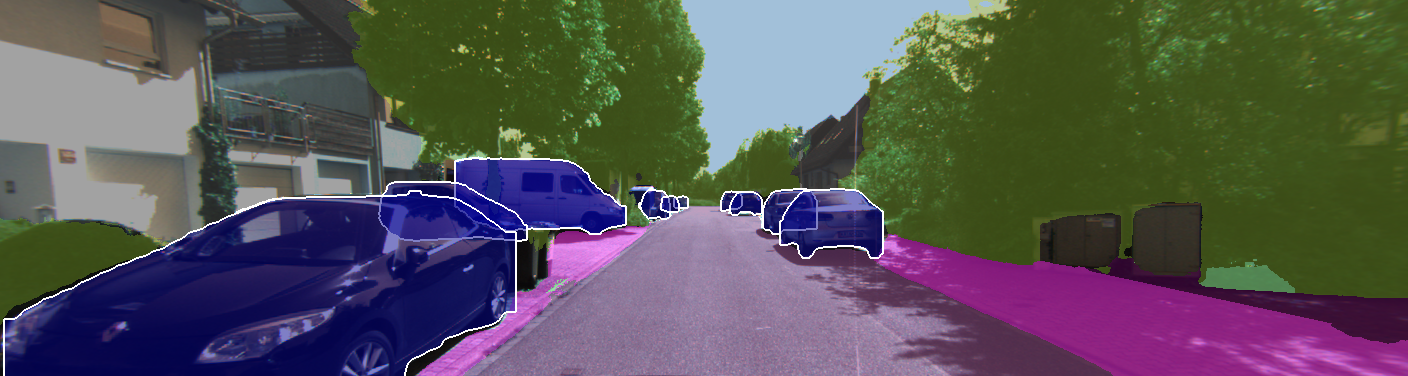}}\\ 
\\
(e) & (f) \\
\\
\\
 \raisebox{-0.4\height}{\includegraphics[width=\linewidth]{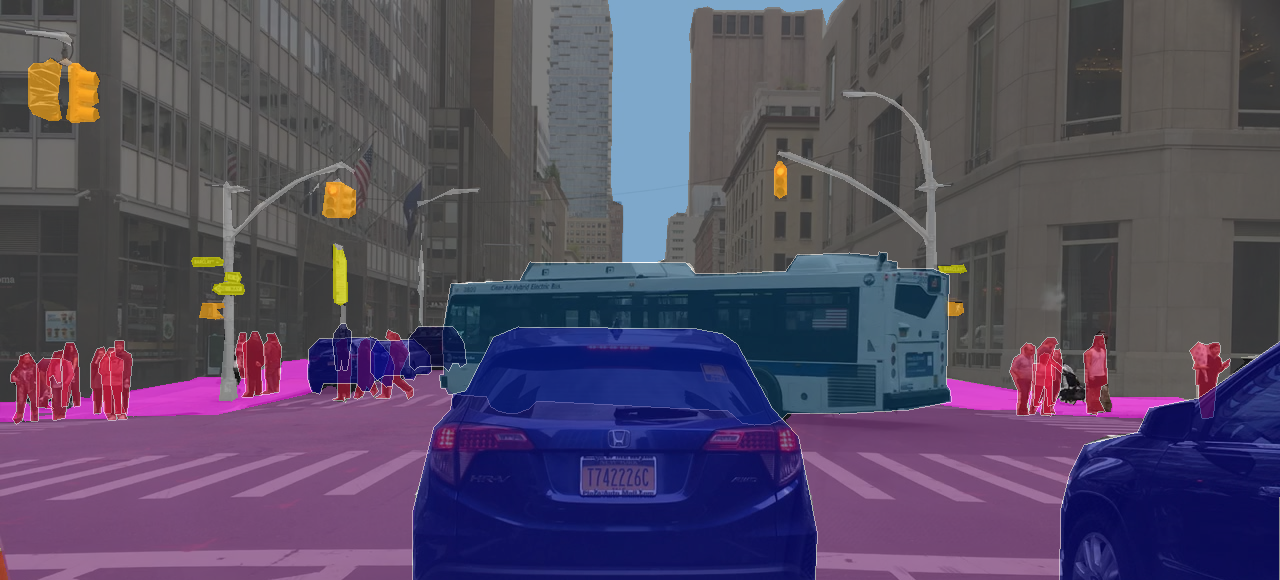}} & \raisebox{-0.4\height}{\includegraphics[width=\linewidth]{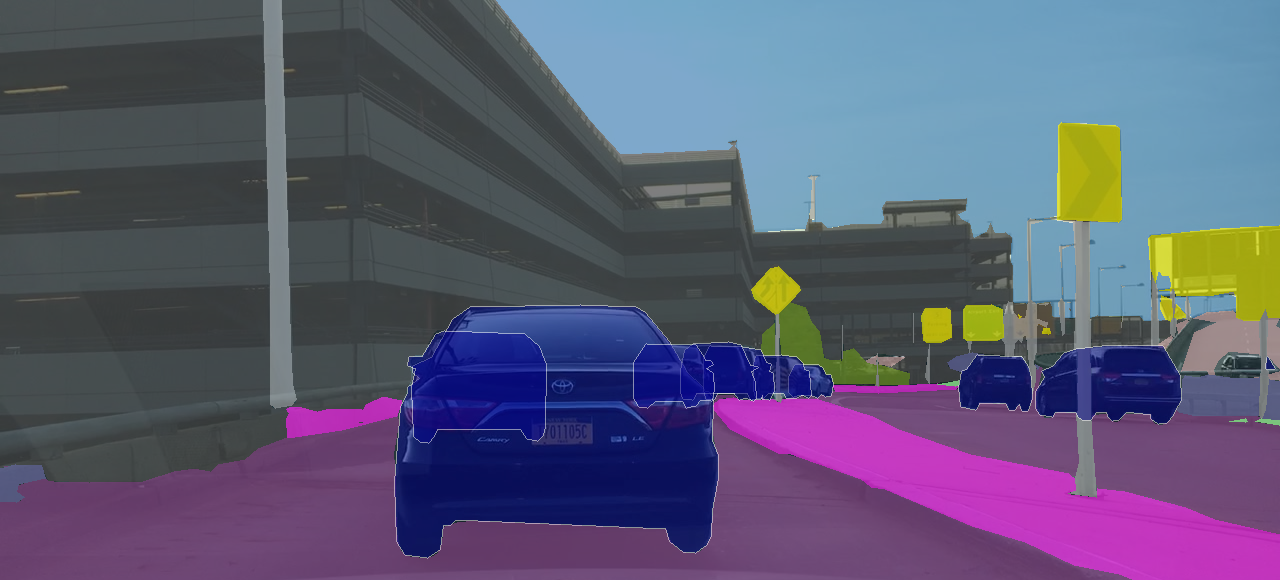}}\\ 
\\
(g) & (h) \\
\\
 \raisebox{-0.4\height}{\includegraphics[width=\linewidth]{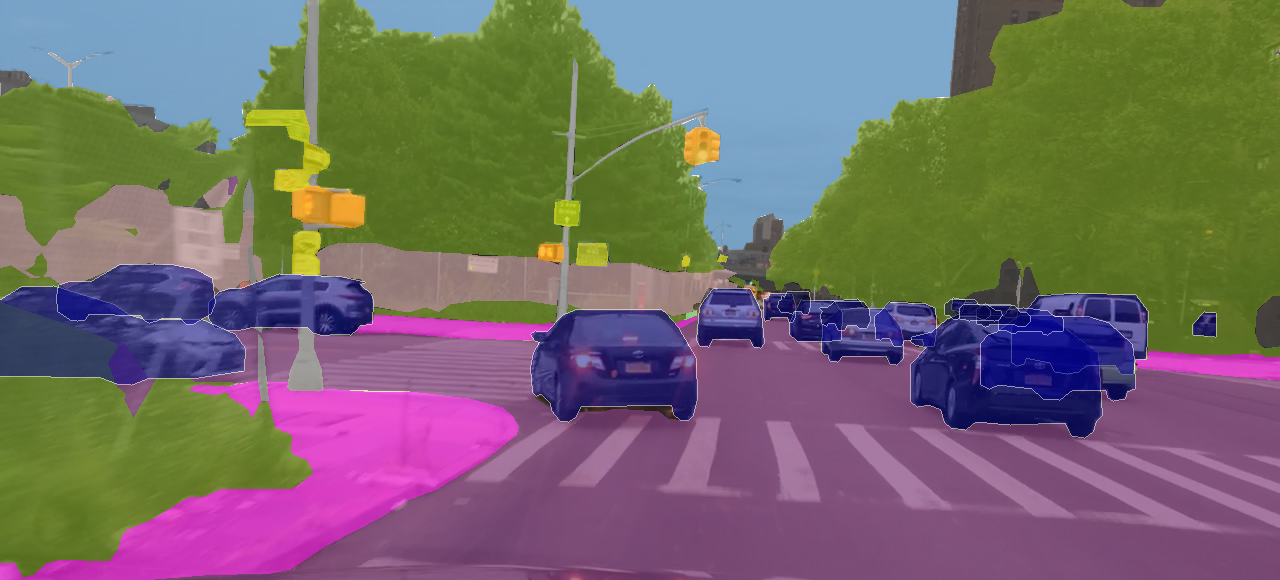}} & \raisebox{-0.4\height}{\includegraphics[width=\linewidth]{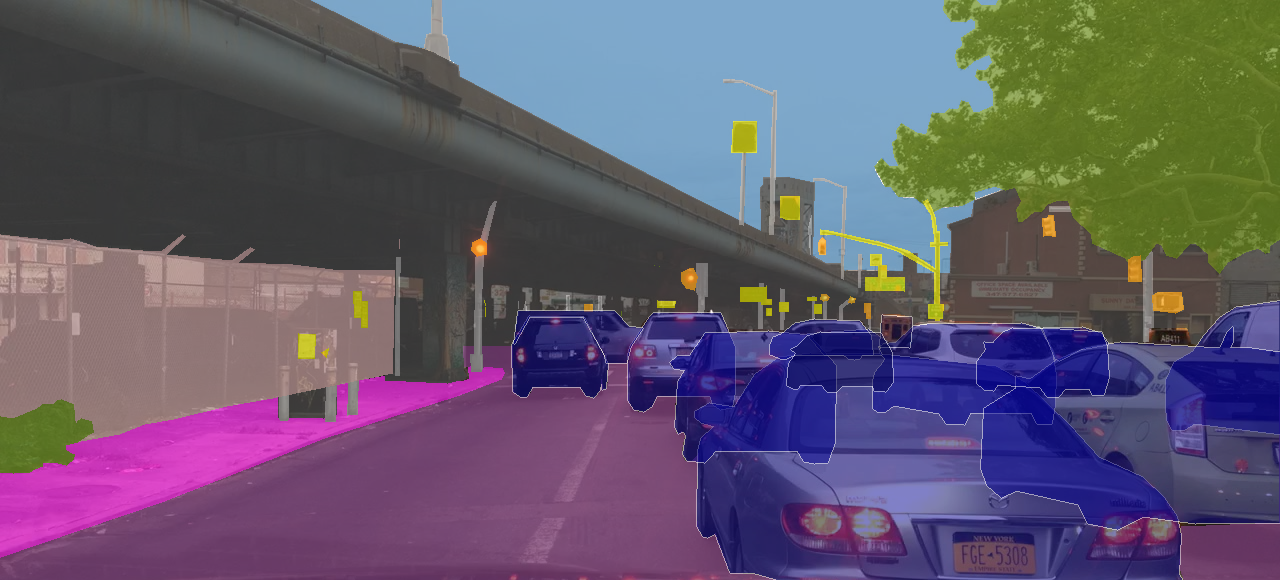}}\\ 
\\
(i) & (j) \\
\\
 \raisebox{-0.4\height}{\includegraphics[width=\linewidth]{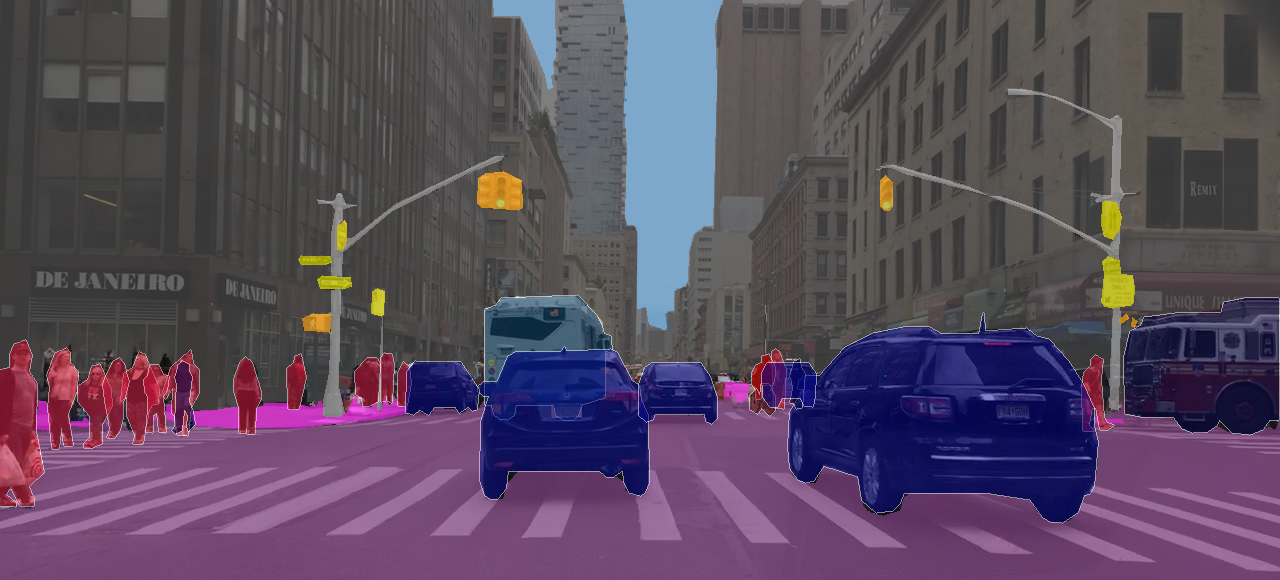}} & \raisebox{-0.4\height}{\includegraphics[width=\linewidth]{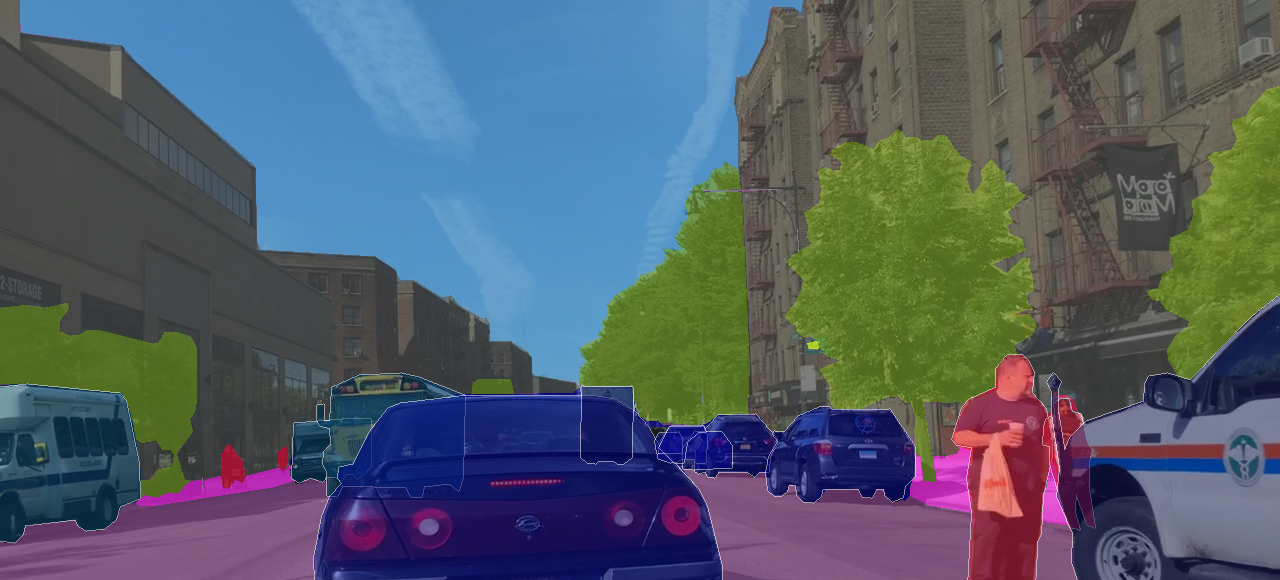}}\\ 
\\
(k) & (l) \\
\\

\end{tabular}}
\vspace{-0.3cm}
\caption{Visualization of amodal panoptic segmentation groundtruth from our proposed KITTI-360-APS (a-f) and BDD100K-APS (g-l) datasets. In (a) and (f) the second car on the left, (e) the far away cars on the left are heavily occluded by other car instances and vegetation, respectively. Similarly, in (h) and (l) the center cars occlude the car and the truck in front of them to a high degree, respectively. Moreover, we also observe a varying degree of occlusion from partial to mid in all of the visualization examples. The variations in occlusion of instances, cluttered urban road scenes with several \textit{thing} class instances, and complex \textit{stuff} classes makes both the proposed datasets extremely challenging for amodal panoptic segmentation.} 
\label{fig:visual_gt}
\vspace{-0.2cm}
\end{figure*}

\begin{figure*}
    \centering
    \begin{subfigure}[b]{0.20\linewidth}
        \centering
        \includegraphics[width=\textwidth]{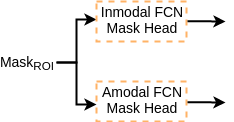}
        \subcaption{Amodal-EfficientPS}
        \label{amodal_eps}
    \end{subfigure}
        \hfill
    \begin{subfigure}[b]{0.3\linewidth}
        \centering
        \includegraphics[width=\textwidth]{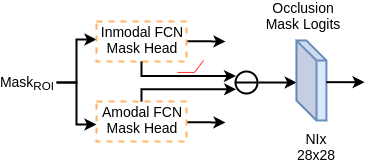}
        \subcaption{ORCNN}
        \label{orcnn}
    \end{subfigure}
    \hfill
        \begin{subfigure}[b]{0.3\linewidth}
        \centering
        \includegraphics[width=\textwidth]{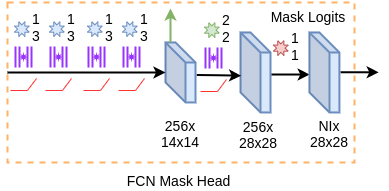}
    \end{subfigure}
   \\
   \hfill
   \\
    \begin{subfigure}[b]{0.2\linewidth}
        \centering
        \includegraphics[width=\textwidth]{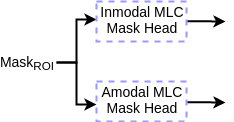}
        \subcaption{ASN}
        \label{ASN}
    \end{subfigure}
    \hfill
    \begin{subfigure}[b]{0.40\linewidth}
        \centering
        \includegraphics[width=\textwidth]{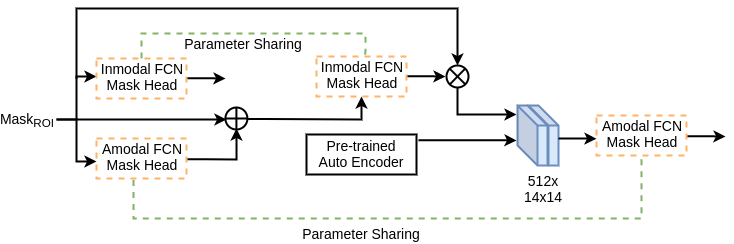}
        \subcaption{Shape Prior}
        \label{shape_prior}
    \end{subfigure}
    \hfill
    \begin{subfigure}[b]{0.35\linewidth}
        \centering
        \includegraphics[width=\textwidth]{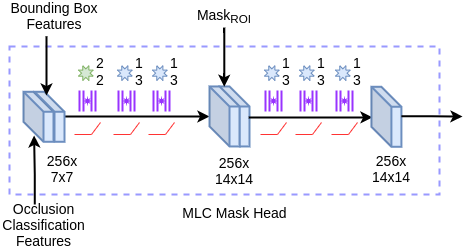}
    \end{subfigure}
   \\
   \hfill
   \\
    \begin{subfigure}[b]{0.5\linewidth}
        \centering
        \includegraphics[width=\textwidth]{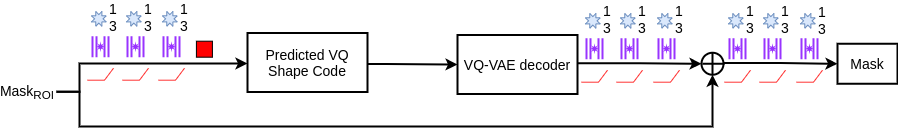}
        \subcaption{VQ-VAE}
        \label{vqvae}
    \end{subfigure}
   \\
   \hfill
   \\
    \begin{subfigure}[b]{0.3\linewidth}
        \centering
        \includegraphics[width=\textwidth]{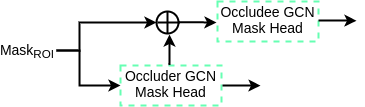}
        \subcaption{BCNet}
        \label{bcnet}
    \end{subfigure}
    \hspace{0.2cm} 
    \begin{subfigure}[b]{0.3\linewidth}
        \centering
        \includegraphics[width=\textwidth]{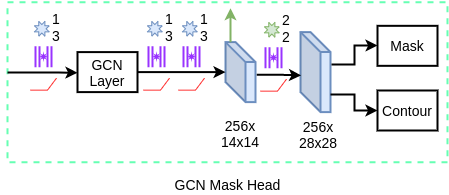}
    \end{subfigure}
    \\
    \vspace{0.2cm}
    \begin{subfigure}[b]{0.7\linewidth}
        \centering
        \includegraphics[width=\textwidth]{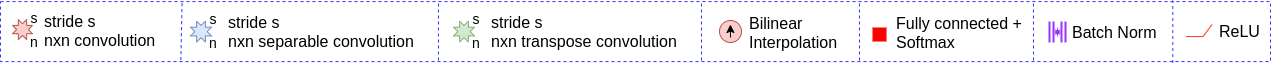}
        \label{panoptic_segmentation_egs}
    \end{subfigure}
\vspace{-0.5cm}
    \caption{Topologies of various amodal instance segmentation head of the amodal panoptic segmentation baselines. Please note that the boxes enclosed in color dashes in each of the architecture corresponds to the expanded version of the same colored boxes depicted on the right.}
    \label{fig:bnetwork}
\end{figure*}

\section{Baseline Architectures}
\label{sec:supp_baseline}

We introduce a total of six baselines for our proposed amodal panoptic segmentation task. We create the baselines by building upon the EfficientPS~\cite{mohan2020efficientps} model which is a state-of-the-art top-down panoptic segmentation network. The EfficientPS architecture consists of four parts. The first part is the shared backbone which is a combination of an encoder and a feature pyramid network (FPN) variant. We employ the EfficientNet-B5~\cite{tan2019efficientnet} model as the encoder and remove its squeeze and excitation~\cite{hu2018squeeze} connections. We also replace the batch normalization and activation layers with synchronized Inplace Activated Batch Normalization (iABN sync)~\cite{bulo2018place} and Leaky ReLU activations respectively. The backbone uses the 2-way FPN~\cite{mohan2020efficientps} on top of the encoder to bidirectionally aggregate multi-scale features. The encoded multi-scale features from the backbone are then propagated to an instance and semantic head. The instance head is a variant of Mask R-CNN~\cite{he2017mask} where the convolution operation in the mask prediction heads is replaced by depth-wise separable convolutions. The semantic segmentation head incorporates various modules to focus on modeling of different feature representations: DPC~\cite{chen2018searching} for capturing long-range contextual information, LSFE~\cite{mohan2020efficientps} for capturing characteristic features, and MC~\cite{mohan2020efficientps} for aligning mismatched correction modules. The final component of EfficientPS is an adaptive fusion module that fuses the output of instance and semantic head based on their logits.

In the baseline architectures, we keep all the components of EfficientPS intact except for the instance segmentation head which is replaced by different state-of-the-art amodal instance segmentation heads namely, Amodal EfficientPS, ORCNN~\cite{follmann2019learning}, VQ-VAE~\cite{jang2020learning}, Shape Prior~\cite{yuting2021amodal}, ASN~\cite{qi2019amodal}, and BCNet~\cite{Ke_2021_CVPR}. In the following, we provide a brief overview of the amodal instance segmentation heads of the baselines.

\begin{figure*}
\centering
\includegraphics[width=\linewidth]{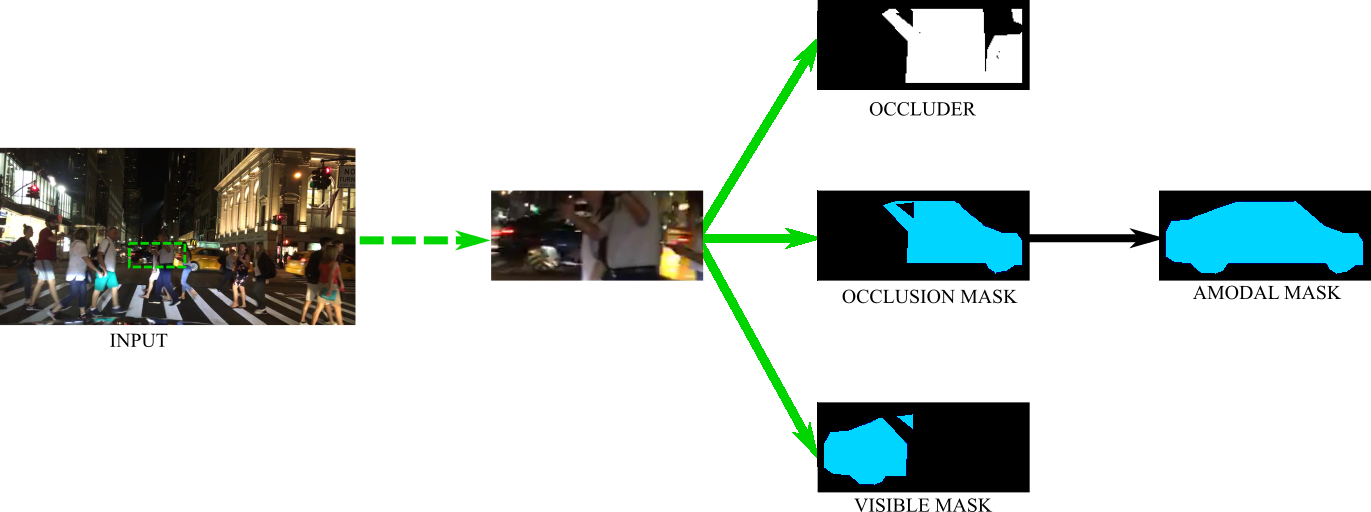}
\caption{Illustration of our fragmentation of bounding box of the target object into class-agnostic occluder, class-wise occlusion, and visible masks. Our amodal instance head employs individual mask heads to predict each mask. The features from these mask heads are further processed with a series of convolution operations along with spatio-channel attention to predict the amodal mask of the target object.}
\label{fig:intromask}
\end{figure*}

\begin{enumerate}[noitemsep]
\item \textbf{Amodal-EfficientPS} is an extension of its inmodal variant and relies implicitly on the network to learn the relationship between the occluder and occludee along with modeling the appropriate class-specific structures.~\figref{fig:bnetwork}~\subref{amodal_eps} presents the amodal instance head of Amodal-EfficientPS.

\item \textbf{ORCNN}~\cite{follmann2019learning} employs an invisible mask prediction head in addition to the inmodal and amodal mask prediction heads, to explicitly learn the propagation from visible mask to amodal mask. To do so, the approach designs the invisible mask prediction by abstracting the amodal mask from the visible mask.~\figref{fig:bnetwork}~\subref{orcnn} shows the amodal instance head of ORCNN.

\item \textbf{ASN}~\cite{qi2019amodal} head emphasizes the importance of global information in addition to visible cues for amodal mask prediction.~\figref{fig:bnetwork}~\subref{ASN} presents the ASN amodal instance head. It consists of an additional occlusion classification branch and uses the features from this branch through a multi-level coding (MLC) block to impart the learned global information to the individual inmodal and amodal mask prediction head. The MLC block essentially takes the concatenation of bounding box features and occlusion features from their respective classification branches, performs a series of transpose convolution-convolution operations to process the collective features, and then concatenates it with the model-specific mask features. This is followed by another series of convolution operations to generate the final modal-specific mask predictions.

\item \textbf{Shape Prior}~\cite{yuting2021amodal} approach strongly supports the idea of using the visible region segmentation in conjunction with shape priors as the key to better amodal mask segmentation.~\figref{fig:bnetwork}~\subref{shape_prior} depicts the amodal instance head of this approach. The aforementioned head employs two modal-specific fully convolutional network heads with parameter sharing. The first modal-specific heads give the initial mask predictions that are further used as attention for refining the final mask predictions with a feature matching loss and pre-trained shape prior autoencoder. Additionally, the approach also incorporates the shape-prior autoencoder in the non-maximum suppression step of the amodal bounding boxes~\cite{ren2015faster}.

\item \textbf{VQ-VAE}~\cite{jang2020learning} seeks to incorporate shape prior information through discrete shape codes while using Vector Quantized Variational Autoencoder for mask segmentation.~\figref{fig:bnetwork}~\subref{vqvae} shows the amodal instance head of VQ-VAE. 

\item \textbf{BCNet}~\cite{Ke_2021_CVPR} models occluder and occludee with a bilayer GCN layer. To be precise, the approach first predicts the occluder mask and contour segmentation and uses these occluder features in conjunction with the ROI features to segment the occludee or the target object in a class agnostic manner.~\figref{fig:bnetwork}~\subref{bcnet} presents the amodal instance head of this approach. In contrast, our APSNet employs FCN based class agnostic occluder mask segmentation head to coarsely model the occlusion regions of the target object as a strong prior and is further refined in a spatially independent manner with an occlusion mask segmentation head. Moreover, we use additional processing blocks with spatio-channel attention to explicitly model the underlying relationship among occluder (general location and shape of the occluded region), occludee (visible region), and the occlusion (precise shape of the occluded region) features before finally computing the amodal mask segmentation. \figref{fig:intromask} illustrates our fragmentation of the amodal bounding box of a target object.

\end{enumerate}

To summarize, for a better amodal perception performance, an amodal instance head should have the ability to decipher the existence of occlusion regions and be able to reason about the shape given the visible region features. We build our APSNet on these two core ideas.

\section{Inference}
\label{sec:supp_inference}

At inference time, to obtain the amodal panoptic segmentation output, we fuse the amodal instance segmentation and the semantic segmentation predictions. There are several fusion heuristics~\cite{kirillov2019panoptic, xiong2019upsnet, mohan2020efficientps} that have been proposed for panoptic segmentation. We adapt the panoptic fusion proposed in~\cite{mohan2020efficientps} due to its superior performance over other fusion approaches. This heuristic allows adaptive fusion of the task-specific head outputs, which can alleviate the inherent overlap problem between the outputs of the different heads. The semantic head generates semantic logits of $|C_{s}|+|C_{t}|$ channels where $C_{s}$ and $C_{t}$ are the set of \textit{stuff} and \textit{thing} semantic classes. While the amodal instance head outputs a set of object instances consisting of a class prediction score, confidence score, amodal bounding box prediction, inmodal, and amodal mask logits. To apply the panoptic fusion, we need to compute two logits $ML_A$ and $ML_B$. We begin with the computation of logit $ML_A$ where we apply confidence thresholding to reduce the number of instances followed by the ROI sampling operation for the amodal bounding box on the two model-specific logits to increase their resolution from $28\times28$ to the input image resolution $H\times W$. Here, $H$ and $W$ are the height and width of the input image. Subsequently, we compute the inmodal bounding box from the inmodal mask derived from the inmodal mask logits. We then sort the class prediction, the modal-specific logits, and the inmodal bounding box according to the class confidence score. We then employ overlap thresholding using the inmodal mask logits to finally yield the mask logit $ML_A$.\looseness=-1

We compute the second mask logit $ML_B$ for the corresponding instances of objects from the semantic head logits by selecting the channel based on the class of the instance and zero-out the logits for that channel outside the inmodel bounding box. Lastly, we fuse the two logits $ML_A$ and $ML_B$ as
\begin{equation}
FL = (\sigma(ML_A) + \sigma(ML_B) ) \odot (ML_A + ML_B),
\end{equation}
where $\sigma (\cdot)$ is the sigmoid function and $\odot$ is the Hadamard product.

We then concatenate the \textit{stuff} logits from the semantic head logits with $FL$. Subsequently, we apply softmax and the argmax operation along the channel dimension to obtain the so-called intermediate prediction ($IP$). In the final step, we zero out the \textit{stuff classes} class labels and copy the semantic head prediction \textit{stuff} labels to the zero places in $IP$.  We obtain the amodal mask for each instance in $IP$ by accessing the amodal mask logits channels according to the instance ID. We then compute the sigmoid of the selected amodal mask logits and threshold it at $0.5$ to obtain the final amodal binary mask. Following, we set the pixels in the amodal binary mask to $2$ that does not overlap with the corresponding instance ID mask in $IP$ to represent its occluded regions. The set of this tensor along with its class prediction and instance ID is concatenated with $IP$ to yield the final amodal panoptic prediction.

\section{Amodal Instance Head}
\label{sec:supp_amodal}

\begin{figure*}
\centering
\footnotesize
\setlength{\tabcolsep}{0.1cm}
{\renewcommand{\arraystretch}{1}
\begin{tabular}{P{0.4cm}P{3.0cm}P{3.0cm}P{3.0cm}P{3.0cm}}
& \raisebox{-0.4\height}{Inmodal Mask} & \raisebox{-0.4\height}{Amodal Mask} & \raisebox{-0.4\height}{Spatially Dependent} & \raisebox{-0.4\height}{Spatially Independent} \\
& \raisebox{-0.4\height}{} & \raisebox{-0.4\height}{} & \raisebox{-0.4\height}{Occlusion Mask} & \raisebox{-0.4\height}{Occlusion Mask} \\
\\
\rot{(a)} & \raisebox{-0.4\height}{{\includegraphics[frame, width=0.5\linewidth]{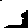}}} & \raisebox{-0.4\height}{{\includegraphics[frame, width=0.5\linewidth]{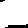}}} & \raisebox{-0.4\height}{{\includegraphics[frame, width=0.5\linewidth]{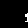}}} & \raisebox{-0.4\height}{{\includegraphics[frame, width=0.5\linewidth]{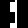}}} \\
\\
\rot{(b)} & \raisebox{-0.4\height}{{\includegraphics[frame, width=0.5\linewidth]{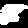}}} & \raisebox{-0.4\height}{{\includegraphics[frame, width=0.5\linewidth]{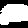}}} & \raisebox{-0.4\height}{{\includegraphics[frame, width=0.5\linewidth]{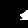}}} & \raisebox{-0.4\height}{{\includegraphics[frame, width=0.5\linewidth]{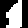}}} \\
\\
\rot{(c)} & \raisebox{-0.4\height}{{\includegraphics[frame, width=0.5\linewidth]{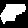}}} & \raisebox{-0.4\height}{{\includegraphics[frame, width=0.5\linewidth]{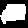}}} & \raisebox{-0.4\height}{{\includegraphics[frame, width=0.5\linewidth]{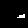}}} & \raisebox{-0.4\height}{{\includegraphics[frame, width=0.5\linewidth]{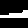}}} \\
\\
\rot{(d)} & \raisebox{-0.4\height}{{\includegraphics[frame, width=0.5\linewidth]{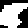}}} & \raisebox{-0.4\height}{{\includegraphics[frame, width=0.5\linewidth]{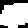}}} & \raisebox{-0.4\height}{{\includegraphics[frame, width=0.5\linewidth]{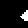}}} & \raisebox{-0.4\height}{{\includegraphics[frame, width=0.5\linewidth, frame]{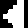}}} \\
\\
\end{tabular}}
\caption{Illustration of spatially dependent and independent occlusion masks. The spatially dependent occlusion mask consists of few pixels compared to the inmodal mask for partial occlusion. On the other hand, the spatially independent occlusion masks that effectively capture the underlying shape of the occluded regions are denser. Thus, enabling stronger feedback during training and consequently resulting in capturing the underlying shape of the occlusion mask effectively.}
\label{fig:occ_mask_egg}
\end{figure*}

To recapitulate, our proposed amodal segmentation head aims to impart the awareness of the presence of occlusion regions with coarse localization (occluder head) and learn to perceive the occlusion shape given the visible and occluder regions. It also models the necessary interconnecting features of occlusion, occluder, and visible regions (processing block with spatio-channel attention) to be able to predict the amodal mask. Additionally, it uses the computed amodal features to further refine the inmodal mask prediction. Further, to efficiently train the occlusion mask head with dense feedback, our APSNet opts to learn spatially independent occlusion masks. \figref{fig:occ_mask_egg} presents examples of the spatially dependent and independent occlusion groundtruth masks.

The amodal instance segmentation head of APSNet consists of object classification, bounding box regression, and various mask heads.  
The training loss for bounding box object classification head ${L}_{cls}$ and the bounding box regression head ${L}_{bbx}$ is the same as defined in \cite{mohan2020efficientps}. Similarly, the visible mask head loss ${L}_{mask}^{v}$, occluder mask head loss ${L}_{mask}^{od}$, occlusion mask head loss ${L}_{mask}^{ol}$, amodal mask head loss ${L}_{mask}^{am}$ and inmodal mask head loss ${L}_{mask}^{inm}$ are akin to ${L}_{mask}$ in \cite{mohan2020efficientps} given as
\begin{equation}
{L}_{mask}(\Theta) = -\frac{1}{|K_p|}\sum_{(P,P')\in K_p}L_{p}(P,P'),
\label{eq:mask}
\end{equation}
where $L_{p}(P,P')$ is the binary cross-entropy loss, $P$ is the ground truth binary mask, $P'$ is the predicted binary mask and $K_p$ is the set of positive matches.

Thus the overall loss for our proposed amodal instance segmentation head is as
\begin{equation}\label{eq:ourmask}
\begin{split}
{L}_{ainst} = {L}_{cls} + {L}_{bbx_{am}} + {L}_{bbx} +  {L}_{mask}^{v}\; +\\ {L}_{mask}^{od} + 
{L}_{mask}^{ol} + {L}_{mask}^{am} + {L}_{mask}^{inm}.
\end{split}
\end{equation}

Note that the gradient from the loss ${L}_{ainst}$ does not flow through the RPN.

\section{Extended Benchmarking Results}
\label{sec:supp_benchmark}
\begin{table*}
\footnotesize 
\centering
\begin{tabular}{p{2.8cm}|p{0.4cm}p{0.4cm}p{0.5cm}p{0.5cm}p{0.5cm}p{0.5cm}p{0.5cm}p{0.5cm}p{0.5cm}p{0.5cm}p{0.3cm}p{0.6cm}}
\toprule
 &  APQ  & APC & APQ$_S$ &APQ$_T$ & APQ$^V_T$ & APQ$^O_T$ & APC$_S$&  APC$_T$ & APC$^V_T$ & APC$^O_T$ & AP & mIOU\\
\midrule
Amodal-EfficientPS & $41.1$ & $57.6$ & $46.2$ & $33.1$ & $41.3$ & $12.7$ & $58.1$ & $56.6$ & $58.5$ & $22.7$ & $29.1$ & $44.7$ \\
ORCNN~\cite{follmann2019learning}  & $41.1$ & $57.5$ & $46.2$ & $33.1$ & $41.1$ & $12.8$ & $58.1$ & $56.6$ & $58.1$ & $22.9$ & $29.0$ & $44.5$  \\
BCNet~\cite{Ke_2021_CVPR}  & $41.6$ & $57.9$ & $46.2$ & $34.4$ & $42.0$ & $14.5$ & $58.1$ & $57.6$ & $59.7$ & $23.9$ & $30.3$  & $45.8$  \\
VQ-VAE~\cite{jang2020learning}  & $41.7$ & $58.0$  & $46.2$ & $34.6$ & $42.2$ & $14.7$ & $58.1$ & $57.8$ & $59.8$ & $23.9$ & $30.4$ & $45.9$   \\
Shape Prior~\cite{yuting2021amodal}  & $41.8$ & $58.2$  & $46.2$ & $35.0$ & $42.5$ & $15.3$ & $58.1$ & $58.2$ & $60.3$ & $24.3$ & $31.0$ & $46.3$ \\
ASN~\cite{qi2019amodal} & $41.9$ & $58.2$ & $46.2$ & $35.2$ & $42.7$ & $15.4$ & $58.1$ & $58.3$ & $60.4$ & $24.2$ & $31.1$ & $46.3$    \\
\midrule
\mbox{APSNet} (Ours)  &  $\mathbf{42.9}$ & $\mathbf{59.0}$  & $\mathbf{46.7}$ & $\mathbf{36.9}$ &$\mathbf{43.6}$ & $\mathbf{18.3}$  & $\mathbf{58.5}$ & $\mathbf{59.9}$  &$\mathbf{61.5}$ & $\mathbf{25.8}$& $\mathbf{33.4}$ &$\mathbf{48.0}$ \\
\bottomrule
\end{tabular}
\caption{Performance comparison of amodal panoptic segmentation on the KITTI-360-APS validation set. Subscripts $S$ and $T$ refer to \textit{stuff} and \textit{thing} classes respectively. Subscripts $S$ and $T$ refer to \textit{stuff} and \textit{thing} classes respectively. Superscripts $V$ and $O$ refer to visible and occluded regions respectively. All scores are in [\%].}
\label{tab:kittiEvaluations}
\end{table*}

In this section, we discuss the benchmarking results on the KITTI-360-APS validation set in detail to understand the mutual relationship between the two proposed metrics clearly. \tabref{tab:kittiEvaluations} presents the quantitative results using the APQ and APC metrics and all their components. The APS baseline with the trivial implementation of amodal instance head, Amodal-EfficientPS achieves the lowest APQ and APC scores. Similarly, ORCNN that employs a derivative head for occlusion mask prediction over the trivial amodal instance head attains similar performance to Amodal-EfficientPS. However, this similar overall performance of the two networks stems from varying effectiveness of segmenting the visible and invisible regions rather than being the same. APS-EfficientPS has higher APQ$^V_T$ and APC$^V_T$  scores implying better visible \textit{thing} region parsing, whereas ORCNN has higher APQ$^O_T$ and APC$^O_T$  values, indicating better occluded \textit{thing} region parsing. Following, BCNet performs better than Amodal-EfficientPS and ORCNN, lagging behind VQ-VAE by $0.1\%$ in both APQ and APC scores. However, BCNet achieves an improvement of $1.9\%$ in APQ$^O_T$ and $1.2\%$ in APC$^O_T$ scores. This difference in the proportional improvement in the two metrics where the increase in performance is higher for APQ$^O_T$ signifies that BCNet primarily improves the segmentation of partially occluded objects with smaller occlusion regions. When paired with the improvement in APQ$^V_T$ of $0.7\%$ and APC$^V_T$ of $1.2\%$ indicates that the approach improves the segmentation of nearby larger objects that are partially occluded. We hypothesize that this is primarily due to the bilayer modeling of occluder and occludee which enables more refined target mask segmentation.

Subsequently, VQ-VAE adds an occlusion detection branch and mask refinement with shape priors to incorporate amodal reasoning capabilities. Compared to the trivial Amodal-EfficientPS, this approach achieves an improvement of $0.7\%$ in APQ and $0.4\%$ in APC, where the improvement in APQ$_T$ and APC$_T$ component of the metrics is $1.5\%$ and $1.2\%$ respectively. Next, the Shape Prior model refines the coarsely predicted mask with shape priors in addition but uses a combination of a pre-trained autoencoder with K-Means based codebook. It further incorporates a visible mask refinement step with amodal features. This model has an APQ score of $41.8\%$ and an APC score of $58.2\%$. Its APQ$^O_T$ and APC$^O_T$ are higher than that of VQ-VAE by $0.6\%$ and $0.4\%$ respectively. This improvement suggests that incorporating shape priors with an additional codebook yields better performance. A similar trend is also observed for visible \textit{thing} region metrics indicating that refining visible mask with amodal features helps improve the performance further.

Nevertheless, our proposed approach performs the best in all the metrics, namely APQ and APC, and their components. Here, the proportional improvement of APSNet can be observed in visible and occlusion components of APQ${_T}$ ($0.9\%$ and $2.9\%$) and APC${_T}$ ($0.9\%$ and $1.6\%$) compared to the best baselines ASN. This demonstrates that our approach improves the segmentation of partial-to-mid occluded objects, however, the performance is limited when it comes to heavily occluded objects, as observed in the qualitative evaluations in the manuscript. To conclude, computing both the metrics for the amodal panoptic segmentation task gives more insights into the performance of an approach, which can be extremely valuable while developing an effective solution for this problem.






\end{document}